\documentclass[twocolumn]{article}

\usepackage{arxiv}

\usepackage[utf8]{inputenc} 
\usepackage[T1]{fontenc}    
\usepackage{hyperref}       
\usepackage{url}            
\usepackage{booktabs}       
\usepackage{amsfonts}       
\usepackage{amsmath}
\usepackage{nicefrac}       
\usepackage{microtype}      
\usepackage{cleveref}       
\usepackage{lipsum}         
\usepackage{graphicx}
\usepackage[numbers]{natbib}
\usepackage{doi}
\usepackage{capt-of}
\usepackage{algorithm}
\usepackage{algpseudocode}
\usepackage{amssymb}
\usepackage{subfigure} 
\newtheorem{definition}{Definition}
\newtheorem{theorem}{Theorem}
\newtheorem{proposition}{Proposition}

\title{CoSIFL: Collaborative Secure and Incentivized Federated Learning with Differential Privacy}

\date{}

\newif\ifuniqueAffiliation
\uniqueAffiliationtrue

\ifuniqueAffiliation 
\author{
	Zhanhong Xie \footnotemark[1] \\ 
	\texttt{xzh.@e.gzhu.edu.cn}
	\And
	Meifan Zhang \footnotemark[1] \footnotemark[2] \\ 
	\texttt{zhangmf@gzhu.edu.cn}
	\And
	Lihua Yin \footnotemark[1] \\  
	\texttt{yinlh@gzhu.edu.cn}\\	
}
\else
\author[1]{Zhanhong Xie}
\author[1]{Meifan Zhang\thanks{Corresponding author: \texttt{zhangmf@gzhu.edu.cn}}}
\author[1]{Lihua Yin}
\affil[1]{XXX University}
\footnotetext[1]{All authors are with XXX University.}
\fi

\renewcommand{\shorttitle}{\textit{arXiv} Template}
\renewcommand{\undertitle}{}

\hypersetup{
pdftitle={A template for the arxiv style},
pdfsubject={q-bio.NC, q-bio.QM},
pdfkeywords={Federated Learning , Differential Privacy , Game Theory},
pdfborder={0 0 0}
}

\begin{document}
\twocolumn[{
  \begin{@twocolumnfalse}

	\maketitle

	\vspace{1em}

	\begin{abstract}
		Federated learning (FL) has emerged as a promising paradigm for collaborative model training while preserving data locality. However, it still faces challenges from malicious or compromised clients, as well as difficulties in incentivizing participants to contribute high-quality data under strict privacy requirements. Motivated by these considerations, we propose CoSIFL, a novel framework that integrates proactive alarming for robust security and local differential privacy (LDP) for inference attacks, together with a Stackelberg-based incentive scheme to encourage client participation and data sharing. Specifically, CoSIFL uses an active alarming mechanism and robust aggregation to defend against Byzantine and inference attacks, while a Tullock contest–inspired incentive module rewards honest clients for both data contributions and reliable alarm triggers. We formulate the interplay between the server and clients as a two-stage game: in the first stage, the server determines total rewards, selects participants, and fixes global iteration settings, whereas in the second stage, each client decides its mini-batch size, privacy noise scale, and alerting strategy. We prove that the server-client game admits a unique equilibrium, and analyze how clients’ multi-dimensional attributes—such as non-IID degrees and privacy budgets—jointly affect system efficiency. Experimental results on standard benchmarks demonstrate that CoSIFL outperforms state-of-the-art solutions in improving model robustness and reducing total server costs, highlighting the effectiveness of our integrated design.
	\end{abstract}

  \end{@twocolumnfalse}
  \vspace{1em}
}]

\footnotetext[1]{Cyberspace Institute of Advanced Technology, Guangzhou University, Guangzhou, 510006, China}
\footnotetext[2]{Meifan Zhang is the corresponding author.}

\section{Introduction}\label{sec:introduction}
Federated learning~\cite{01,02} is a collaborative learning technique that enables multiple participants to jointly train a machine learning (ML) model without sharing their respective training data. Each participant trains the model locally and then uploads the local updates to a central server. The server aggregates these local updates to generate a global model, which is subsequently distributed back to all participants. FL has two key characteristics:  
1) It enables training with massive datasets from thousands of participants without revealing their private data, thus preserving privacy.  
2) Its decentralized nature allows computations to run concurrently across multiple devices, significantly enhancing training speed and alleviating computational burdens on the server.

Benefiting from these advantages, FL has been applied across various fields such as training language models~\cite{03}, health monitoring~\cite{04}, autonomous driving~\cite{05}, Industry 4.0~\cite{06}, and healthcare~\cite{07}. With the expanding applications of FL systems, addressing security and incentive-related challenges has become increasingly urgent.
To fully unleash the potential of FL in real-world applications, however, several critical issues must be addressed. One central concern is the presence of malicious or compromised participants who can corrupt model updates or steal private data, thereby undermining the integrity and trustworthiness of the collaborative process. Another pressing need is to devise effective incentive mechanisms that encourage clients to contribute high-quality data and resources, especially given the substantial costs and privacy risks they may incur. In the following, we outline two major categories of challenges—security and privacy, as well as collaboration and incentives.

\textbf{Challenge 1: Security and privacy.}
The primary security risks stem from potential malicious clients whose goal is to disrupt the learning process. These malicious clients may attack FL systems by uploading local updates trained on poisoned data~\cite{08} or by strategically poisoning local models~\cite{09}. Recently, two types of attacks have garnered significant attention: poisoning attacks aimed at corrupting the global model and inference attacks targeting the theft of local participants’ data. Global model poisoning attacks are executed by malicious participants uploading harmful updates to undermine the global model, encompassing two attack types: untargeted poisoning attacks and targeted model poisoning attacks.
In untargeted poisoning attacks, attackers aim to degrade the global model’s performance on the main task, employing methods such as sign-flipping attacks~\cite{10} or label-flipping attacks~\cite{13}. In targeted model poisoning attacks~\cite{14}, attackers seek to impair the global model’s performance on specific classes while maintaining high accuracy on unaffected samples, thereby enhancing the attack’s stealthiness~\cite{11}. Both types target the server and are collectively categorized as Byzantine attacks. In contrast, inference attacks~\cite{12,34} attempt to reconstruct a client’s training data from individual model updates through gradient inversion. To defend against global model poisoning attacks, existing methods (e.g., Siren+~\cite{13}) enhance security through proactive alerting mechanisms and DP, while FLShield~\cite{14} uses loss impact per class (LIPC) to identify and exclude malicious clients and validators. To counter inference attacks, prior work~\cite{35,36,42,43,44} frequently employs DP to protect gradients.
However, most current approaches concentrate solely on detecting and filtering malicious updates, failing to adequately address the critical need to incentivize benign clients to actively contribute high-quality data in adversarial environments—where malicious users may be present—while still upholding privacy and security.

\textbf{Challenge 2: Collaboration and Incentives. }
Participating clients in federated learning incur substantial costs across multiple resource dimensions (computation and memory). Due to privacy concerns and resource consumption, clients are typically reluctant to participate in FL training without sufficient economic compensation. Therefore, the server must incentivize clients to contribute their data and resources while minimizing overall costs, including recruitment rewards and loss in model accuracy. Simultaneously, without adequate incentive payments and guarantes controllable privacy leakage, clients would be unwilling to actively participate. Thus, facing clients' privacy demands and resource expenditures, the server must devise an effective incentive mechanism to model client interactions, encourage participation, and determine collaboration strategies to enhance training efficiency~\cite{15}. Although existing studies based on Stackelberg game incentives~\cite{16,45,46,47} have attempted to balance client privacy needs with server payments, they often overlook the adverse effects of malicious attacks on model integrity and fail to sufficiently incentivize benign clients to actively contribute high-quality data in adversarial environments—where malicious users may be present—while maintaining privacy and security.

The difficulty of simultaneously addressing these two challenges lies in the inherent trade-offs and conflicting requirements between robust security measures and effective incentive mechanisms. 
On one hand, to safeguard the global model from poisoning and inference attacks, stringent security measures—such as heavy noise injection via DP and proactive alerting—must be employed. These measures, while enhancing security, inevitably degrade the quality of the local model updates by obscuring the true signal. As a result, the model’s convergence performance may suffer, and the accuracy of client contributions can become harder to assess.
On the other hand, the incentive mechanism is designed to motivate clients to actively participate by rewarding them based on their data contributions and the accuracy of their alarms. However, if the security measures (DP noise) are too aggressive, they distort the utility signals on which the incentive mechanism relies, potentially leading to suboptimal client selection and reward allocation. Moreover, clients may face increased resource costs and privacy risks, which further complicate their decision-making process.
Thus, designing a unified framework that can achieve both high security and strong incentives is particularly challenging because improving one aspect (security) can negatively impact the other (incentives), and vice versa. This multi-objective problem requires a carefully balanced approach that dynamically adapts to heterogeneous client behaviors and network conditions while maintaining overall model integrity and efficiency.

To address the above challenges, this paper proposes a unified framework called CoSIFL (Collaborative Secure and Incentivized Federated Learning with Differential Privacy). CoSIFL integrates proactive alerting and LDP for robust defense, along with a Stackelberg game-based incentive mechanism for collaborative participation, providing dual guarantees of security and collaboration. On the security front, CoSIFL proactively detects suspicious updates through alerting and reduces inference attack risks via LDP, enabling robust aggregation and filtering on the server. Regarding incentives, CoSIFL employs a Tullock contest-based reward allocation scheme \cite{17}, allowing clients to receive rewards proportionate to their data contribution and accuracy of alarms. This incentivizes clients to select batch sizes and noise injection strategies beneficial to the overall system.

\begin{figure}
	\centering
	\includegraphics[width=0.53\textwidth]{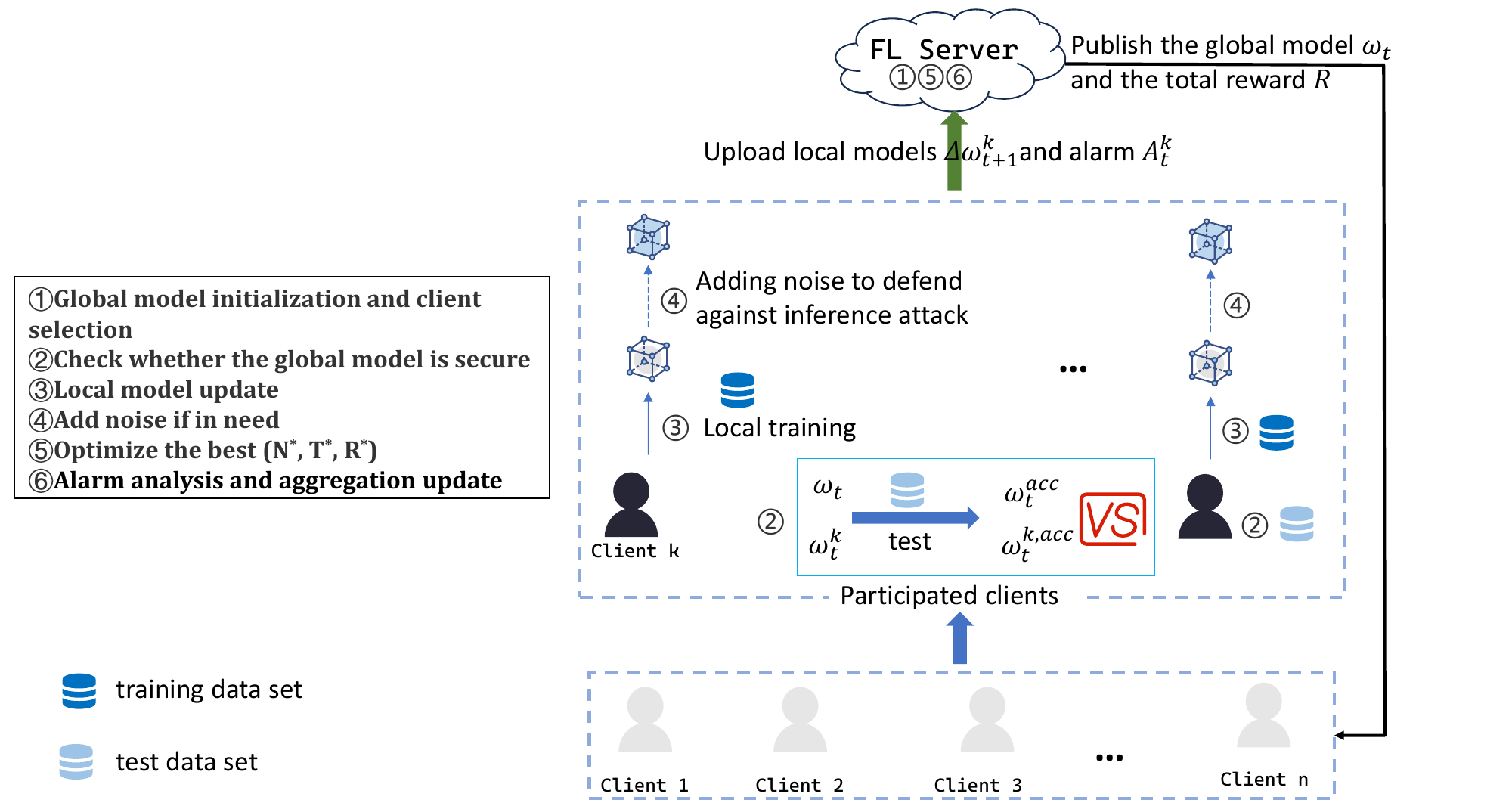}
	\caption{Overview of CoSIFL.}
	\label{Overview of CoSIFL}
\end{figure}
As shown in Figure~\ref{Overview of CoSIFL} (the description of common symbols used throughout the paper is presented in Table~\ref{Descriptions of some Frequent Notations}),  
\normalsize{\textcircled{\scriptsize{1}}}\normalsize the server first initializes the global model \({\omega_{t}}\) and selects which clients will participate in training based on the total reward \(R\).  
\normalsize{\textcircled{\scriptsize{2}}}\normalsize Before local training, the selected client uses a local test dataset \(D_{test}^k\) to evaluate both the global model and the local model, comparing the resulting test accuracies (\(\omega_t^{acc}, \omega_t^{k,acc}\)) to determine whether the global model has been subjected to a poisoning attack, and subsequently sends an alarm message \(A_t^k\) to the server.
\normalsize{\textcircled{\scriptsize{3}}}\normalsize Each client trains the model using their local training dataset \({D_{train}^k}\).  
\normalsize{\textcircled{\scriptsize{4}}}\normalsize To defend against inference attacks, DP noise is applied to protect gradients before uploading the local update \(\Delta\omega_{t+1}^k\) to the server.  
\normalsize{\textcircled{\scriptsize{5}}}\normalsize The server runs Approximate Incentive Optimization~\cite{15} to determine the optimal configuration \((N^*, T^*, R^*)\).  
\normalsize{\textcircled{\scriptsize{6}}}\normalsize Finally, the server analyzes the client alarms \(A_t^k\) and updates the global model accordingly based on the selected strategy.

In summary, the main contributions of this paper are threefold:  
1. Designing a FL defense mechanism combining LDP with proactive alerting to counteract Byzantine and inference attacks;  
2. Proposing a Stackelberg game-based incentive model that incorporates data contributions and alarm quality into reward allocation;  
3. Providing theoretical analysis on FL convergence performance and game equilibrium, and conducting extensive experiments on benchmark datasets. Results demonstrate that CoSIFL significantly reduces overall server costs while maintaining model robustness, thus offering a viable path toward secure and incentivized FL for practical deployment.

\section{Background}

\begin{table}[htbp]
	\centering
	\caption{Descriptions of some Frequent Notations}
	\label{table:notation}
	\begin{tabular}{lp{5.6cm}} 
		\toprule[1.5pt]
		\textbf{Symbol} & \textbf{Descriptions} \\
		\midrule[1pt]
		${\omega _t},\omega _t^k$  & The global model of the t-th round and the local model of client k\\
		$A_t^k$  & Alarm information: 0 - Do not alarm; 1 - Alarm.\\
		$D_{test}^k,D_{train}^k$  & Local test dataset and local training dataset.\\
		$\omega _t^{acc},\omega _t^{k,acc}$  & The test results of the global model and the local model using the local test dataset.\\
		$N,T,R$  & Number of participating clients, global model iterations and total reward.
		${N^*,T^*,R^*}$ is the best choice of $N,T,R$ for FL server\\
		${\alpha _k},{\varepsilon _k},{\gamma _k},{s_k},{t_k}$  & Non-iid degree, privacy budget, alarm reliability,resource cost and latency of client k.\\
		$S_a,S_s,S_b$  & Alarming set, silent set and benign set.\\
		$C_s$  & Measure the difference between the maximum accuracy rate and the accuracy rate of each alarm client.\\
		$C_c$  & Measures the degree to which the client accepts the differences between global model and local model.\\
		\bottomrule[1.5pt]
		\label{Descriptions of some Frequent Notations}
	\end{tabular}
\end{table}

\subsection{Federated Learning}
Federated Learning trains a shared global model through iterative aggregation of local model updates from multiple client devices, without compromising clients' data privacy. Unlike traditional machine learning paradigms requiring centralized data, FL allows local model training without sharing local datasets. In each training round \( t \in [1,T] \), the server sends the current global model \( \omega_t \) to a subset \( S_t \subset S \) randomly selected from the total set of clients \( S \), with size \( n \). Each client \( k \in S_t \) fine-tunes the global model \( \omega_t \) using its local dataset \( D_k \), and then sends the updated parameters \( \omega_t^k \) back to the server. The server aggregates these updates to form the next global model \( \omega_{t+1} \). The objective of FL is to minimize the following objective function:
\begin{equation}
	\min \sum_{t=1,k=1}^{T,N} p_k F_k(\omega_t^k)
\end{equation}
where \(F_k\) represents the local objective function of client \(k\), \( \omega_t^k \) denotes the model parameters of client \(k\) at round \(t\), and \( p_k \) represents the relative influence (weight) of client \(k\).
\subsection{FL Poisoning Attacks}
\subsubsection{Untargeted Poisoning Attacks}
In untargeted poisoning attacks, malicious participants submit manipulated updates aiming to indiscriminately poison the global model, thereby degrading the main accuracy (MA) of the FL task. The objective of this attack is to minimize the following metric:
$MA = \mathop{\mathbb{E}}\limits_{(x,y)\sim D} [\Pr (\omega(x) = y)]$, where \(D\) is the data distribution corresponding to the learning task, and \(y\) is the true label of input \(x\). Common types of untargeted poisoning attacks include the following three: 
Adaptive Attack~\cite{18}: Malicious clients intentionally customize their optimization to move the global model's parameters opposite to the normal updating direction.		
Label-flipping Attack~\cite{19}: Malicious clients train models using normal images but with flipped labels.
Sign-flipping Attack~\cite{20}: Malicious clients perform training similarly to normal clients to obtain normal gradients, but then multiply these gradients by a negative constant before uploading.

\subsubsection{Targeted Poisoning Attack}
Targeted poisoning attacks aim at impairing the predictive performance only on specific samples, while maintaining high accuracy on other samples, thus making these attacks more covert. For this type of attack, attackers aim to maximize the misclassification rate (MR) on a designated subset of targeted data samples \(D_t \subset D\):
$MR = \mathop{\mathbb{E}}\limits_{(x,y)\sim D_t} [\Pr (\omega(x) \ne y)]$, where \( D_t \) is the target sample subset specified by the attacker. Targeted model poisoning~\cite{21} is a representative targeted poisoning attack that strategically manipulates model parameters to severely degrade accuracy on specific data classes while maintaining normal performance on other classes, making the attack particularly stealthy.

\subsubsection{Inference Attack}
Unlike poisoning attacks that degrade model performance, inference attacks aim to infer sensitive information about client data directly from uploaded model updates or gradients, posing significant privacy threats. Common inference attacks include Membership Inference Attacks (MIA)~\cite{22}, which identify if specific samples were in the training set; Property Inference Attacks~\cite{23}, which reveal statistical features or sensitive attributes; and Input/Label Inference Attacks~\cite{24}, reconstructing exact training inputs or labels. To counter these attacks, DP and secure aggregation techniques~\cite{13,14} are widely employed to protect client data.

\subsection{Differential Privacy}

Differential Privacy (DP) is a rigorous mathematical method for privacy protection~\cite{25}. Specifically, \((\varepsilon,\delta)\)-DP provides a distinguishability bound \(\varepsilon\) for all outputs of two neighboring input datasets \(X\) and \(X'\) that differ by at most one instance, with probability at least \(1-\delta\) (where \(0<\delta<1\)). The formal definition of DP is given as follows:

\begin{definition}
	\textbf{(\(\varepsilon, \delta\))-DP}: 
	A randomized algorithm \(M: D \rightarrow R\) satisfies \((\varepsilon, \delta)\)-DP if and only if, for any two neighboring datasets \(X, X' \in D\) and any subset of outputs \(S \subseteq R\), we have:
	\begin{equation}
		\Pr [M(X) \in S] \le e^\varepsilon \Pr [M(X') \in S] + \delta
	\end{equation}
	where \(\varepsilon\) is the privacy budget, quantifying the trade-off between privacy level and utility of the randomized algorithm \(M\), and \(\delta\) represents the probability that the privacy guarantee might fail.
\end{definition}

\subsection{Incentive Mechanisms in Federated Learning}

In FL, clients must consume computational, storage, and communication resources to participate in training. Without sufficient incentives, clients may be unwilling to join FL tasks or contribute low-quality updates, which can degrade the overall model performance. Therefore, designing an effective incentive mechanism is crucial for maintaining system stability and encouraging high-quality data contributions.

\subsubsection{Objectives of Incentive Mechanisms}

The primary objectives of an incentive mechanism in FL include:
Encouraging more clients to join FL training to improve model generalization.
Rewarding high-quality, low-noise contributions while mitigating the impact of malicious clients.
Allocating rewards fairly while considering clients’ privacy protection needs (DP).
Ensuring effective FL training while reducing the total incentive payments required from the server.

\subsubsection{Existing Incentive Mechanisms}

Current FL incentive mechanisms can be categorized as follows:
Contribution-Based Reward Mechanisms: The server allocates rewards based on the quality of the submitted updates, such as gradient variations or training loss reduction. For example, a Tullock contest-based incentive scheme~\cite{17} can distribute rewards based on the relative contributions of clients.
Game-Theoretic Incentive Models: Methods such as Stackelberg games~\cite{23} and contract theory model the interaction between clients and the server to balance privacy costs and data contributions.
Reputation and Trust-Based Incentives: Methods like FLTrust~\cite{26} compute a trust score for each client based on their historical contributions and adjust incentives accordingly to prevent malicious exploitation of rewards.
\subsubsection{Research Challenges}

Despite advancements in FL incentive mechanisms, several challenges remain:
Trade-off Between Privacy and Incentives: Clients using strict privacy-preserving mechanisms (high-noise DP) may contribute lower-quality updates, requiring the server to balance privacy protection with effective incentives.
Incentive Design Under Adversarial Attacks: Most existing incentive schemes assume honest clients, but in the presence of malicious participants, more robust reward allocation strategies are needed to prevent attackers from manipulating contributions for unfair rewards.
Fairness for Resource-Constrained Devices: In scenarios where clients have varying computational capacities, ensuring fair reward distribution and preventing high-power devices from dominating incentives is a key issue.
To address these challenges, this paper proposes CoSIFL, which integrates proactive alarming, DP, and game-theoretic incentives. This framework enhances the security of FL while optimizing incentives, improving client participation and data quality.

\section{System model and problem description}
\subsection{Collaborative Security and Incentive Setting}

In our proposed CoSIFL framework, both robust security defense and incentive-driven collaboration are integrated into the FL process. The participating entities and corresponding workflow are summarized as follows:

\subsubsection{Participating Entities}
Server (Coordinator). The server is responsible for distributing the FL training task, aggregating global models, and issuing monetary or reputational rewards to encourage client participation. It also sets security policies—such as alarm thresholds and DP configurations—to mitigate malicious activity.
Clients (Distributed Devices). Each client holds its own local dataset, typically non-IID (non-independent and identically distributed), and has unique attributes such as privacy demands (a specific privacy budget), resource constraints, or communication latency. Some clients could be compromised by adversaries or perform poisoning attacks, making robust detection and filtering essential.

\subsubsection{Workflow}

	\textbf{Task announcement.}
	The server first announces the FL task, including the model architecture, incentive rules, and security policies (such as alarm thresholds, acceptable noise levels, etc.). It also informs clients about their potential rewards for contributing data and accurately triggering alarms.
	
	\textbf{Client decision.}
	Each client decides whether to participate in the FL task based on its individual cost, privacy requirements, and other relevant factors. Optionally, clients may report (or implicitly communicate) relevant information to the server regarding their willingness or capabilities (such as resource status, privacy budgets, etc.). Clients may opt out if the anticipated rewards fail to offset their costs or privacy risks.
	
	\textbf{Proactive alarming and local training (Round-based).}  
	In each global iteration, participating clients perform the following operations:  
	Clients first conduct a security check on the global model using their local test dataset \(D_{test}^k\) and then send an alarm signal \(A_t^k\) to the server. Subsequently, they train the model using their local training dataset \({D_{train}^k}\). To defend against inference attacks, clients apply DP noise to protect gradients before uploading their model updates to the server.
	
	\textbf{Server-Side Pareto selection and secure aggregation. } 
	After receiving all local updates and alarm signals, the server performs the following steps:  
	Using Pareto selection principles, it selects the optimal subset of clients \(N^*\), the global iteration count \(T^*\), and the total incentive reward \(R^*\). Based on the received alarm signals and robust aggregation strategies, the server filters out suspicious or malicious updates, aggregates the remaining updates to form a new global model, and proceeds to the next iteration.

Through these steps, CoSIFL simultaneously addresses security threats—such as Byzantine attacks and inference attacks—and incentivizes clients to contribute in a privacy-preserving and resource-efficient manner, thus forming a collaborative security and incentive ecosystem for FL.

\subsection{FL Security Module}

The security component of CoSIFL consists of two main parts: proactive alarming and LDP at the client side, and a two-phase detection and aggregation scheme at the server side. These mechanisms jointly safeguard the training process from malicious attacks such as Byzantine corruptions and inference threats.

\subsubsection{Proactive Alarming and Local Differential Privacy}

Local Testing and Alarm Trigger.
Each client holds a small local test set $D_{test}^k$, which is never shared with the server. After receiving the global model ${\omega _{t-1}}$ from the server in round $t$, the client evaluates it on this test set and compares its performance with either the previous local model’s performance or an expected baseline. If a significant drop or abnormal behavior is detected, the client raises an alarm. 0 indicates no triggering of the alarm, while 1 indicates triggering of the alarm.	

Gradient Clipping and Noise Injection (DP).
Before and during each local training round, the client clips its stochastic gradients to a maximum norm C ($||g|| \le C$) and then adds Gaussian noise to the resulting gradients in order to satisfy LDP. Formally, if the client’s privacy budget is $\varepsilon$, a noise scale $\sigma $ can be set according to the standard Gaussian mechanism. By carefully tuning $\sigma $, clients can balance between model utility and privacy protection.

\subsubsection{Server-Side Approximate Incentive Optimization and Secure Aggregation}

After collecting the model updates (or gradients) and alarm information uploaded by each client, the server performs the following Approximate Incentive Optimization and aggregation process:

	\textbf{Approximate Incentive Optimization. } 
	
	Given limited training time and computational resources, the server needs to select an appropriate subset \(N\) of clients while determining the global iteration count \(T\) and total reward \(R\) to minimize overall server costs. Using the Pareto selection principle, the server balances the trade-off between "more data contribution and higher latency" to select the optimal set of clients. The algorithm iteratively removes clients with the longest response times and compares different Pareto solutions to choose the best client subset and reward allocation, achieving efficient client selection and incentive optimization.
	
	\textbf{Alarm/Silence Cross-Analysis and Global Model Aggregation Update.}
	
	Alarm/Silence Cross-Analysis:  
	The server first divides the participating clients in the current round into "alarm set" and "silence set". Based on the performance differences between the alarming clients and the silent clients, the server filters out updates that are obviously malicious or highly suspicious. There are three types of alarm situations, as detailed in Section~\ref{Server-Side Algorithm}.
	Byzantine-Robust Aggregation:  
	The server aggregates the updates from clients based on the three alarm situations, reducing the potential damage of malicious updates to the global model accuracy. If a client is repeatedly detected as malicious or has an obviously high false alarm rate, the server can temporarily exclude that client from subsequent rounds through a penalty mechanism, saving system resources and improving overall credibility. At the same time, the server uses a reward mechanism to allow the banned clients to have a certain probability of rejoining the training.

Through these processes, CoSIFL tightly integrates LDP (to prevent data leakage and limit the acquisition of malicious information) with proactive alarm detection (to identify and filter poisoned updates), ensuring good model convergence and privacy security even in the presence of malicious clients.

\subsection{Federated Incentives and Game Formulation}

CoSIFL not only defends against adversarial behaviors through the security module, but also encourages client participation by explicitly modeling incentives in a game-theoretic framework. This design incorporates the multi-dimensional attributes of each client into both the reward allocation and security decision-making processes.

\subsubsection{Multi-Dimensional Client Attributes}

We characterize each client \( k \) by multiple interrelated attribute vectors, primarily including:

\textbf{Non-IID degree (\(\alpha_k\)): } 
The non-IID degree describes how heterogeneous the client's local dataset distribution is compared to the global data distribution. Formally, client \( k \)'s non-IID degree is bounded by:$|\nabla F_k(\omega) - \nabla F(\omega)| \leq \lambda_k$, where \( F_k(\omega) \) denotes the local objective function of client \( k \), and \( \lambda_k \) measures the maximum deviation of client \( k \)'s data from the global data distribution~\cite{27}. We normalize the non-IID degree as ${\alpha _k} = 1 - {(\frac{{{\lambda _k}}}{{{\lambda _{\max }}}})^2}$, where ${{\lambda _{\max }}}$ is the maximum degree of non-IID that the server can tolerate.
A higher \(\alpha_k\) implies that client \( k \)'s data distribution is closer to the global distribution, potentially yielding more valuable model contributions.

Privacy Budget ($\varepsilon_k$): Indicates how much noise the client is willing (or able) to inject for DP. A larger $\varepsilon_k$ usually corresponds to weaker noise and thus more accurate updates—though at increased privacy risk. We use a concave quadratic function $v_k$ to characterize its impact on the client's utility, which is defined as ${v_k} = 1 - {(1 - \frac{{{\varepsilon _k}}}{{{\varepsilon _{\max }}}})^2}$.

\textbf{Alarm contribution accuracy (\(\gamma_k\)):}  
Describes the reliability or accuracy of client \( k \)'s proactive alarm signals. A higher \(\gamma_k\) implies that the client provides more reliable alarm information for identifying malicious updates.

\textbf{Data usage cost (\(s_k\)):} 
Represents the unit cost (e.g., computation or energy cost) paid by client \( k \) per training sample. The total local cost for client \( k \) during one global iteration is calculated as \( s_k B_k \), where \( B_k \) is the chosen batch size. Higher data usage costs discourage clients from contributing larger batch sizes unless compensated adequately.

\textbf{Response latency (\(t_k\)):} 
Measures the delay for client \( k \) to perform local training and upload updates. Higher latency negatively impacts the number of feasible global training rounds and may decrease the server’s willingness to select such clients.

By modeling these attributes, the server can effectively incentivize clients based on their respective data quality, privacy preferences, and resource constraints, optimizing the overall FL efficiency and robustness.

\subsubsection{Client utility function and server cost.}
Client Utility Function.To motivate clients effectively, CoSIFL employs a Tullock contest–inspired mechanism to calculate rewards. Under this scheme, a client’s utility depends on both its data contributions and its alarm contribution.
Based on the clients' multi-dimensional attributes, each client's utility function in CoSIFL is formally defined as follows:
\begin{equation}
	{U_k} = \underbrace {\frac{{{\alpha _k}{\gamma _k}{B_k}}}{{\sum\nolimits_{j \in K} {{\alpha _j}{\gamma _j}{B_j}} }}R}_{{\mathop{\rm Re}\nolimits} ward(income)} - \underbrace {{s_k}{B_k}}_{Local\ cost}.
\end{equation}
In the formula above: The term $B_k$ from the classical Tullock contest is extended to consider multi-dimensional client attributes, including:
Non-IID data quality $\alpha_k$, batch size $B_k$, and alarm accuracy/contribution $\gamma_k$.
The reward received by client $k$ depends not only on its data batch size $B_k$, but also on the data quality factor ($\alpha_k$) and alarm contribution factor ($\gamma_k$). Specifically, the higher $\alpha_k$ and $\gamma_k$, the larger the reward the client can earn.
$s_k B_k$ represents the local training cost for client $k$, proportional to batch size $B_k$.
Each client's goal is to maximize its own utility by choosing an appropriate batch size $B_k$, balancing the received rewards against its local resource consumption.

Server Cost. The server aims to minimize a combined metric:
\begin{equation}\label{server's cost}
	{C_{server}} = L({\{ {\alpha _k},{\varepsilon _k},{B_k}\} _{k \in N}},T) + R + RiskPenalty,
\end{equation}
where $L({\{ {\alpha _k},{\varepsilon _k},{B_k}\} _{k \in N}},T)$ characterizes the influence of noise and non-IID degree on model accuracy(See Appendix~\ref{global model loss} for details.); $RiskPenalty = {\delta _1} \cdot (MaliciousCount) + {\delta _2} \cdot (FalseAlarmRate)$, $RiskPenalty$ can represent extra costs if the system detects malicious attacks or repeatedly raises false alarms.

\subsubsection{Game Formulation}

To integrate these attributes into a systematic decision process, we formulate a two-stage game between the server and the clients:

Stage I (Server-Side).
The server, acting as a leader, determines:
The subset of clients $N \subseteq K$ selected for training;
The total number of global iterations $T$;
The total incentive budget $R$;
Security parameters (thresholds for penalizing malicious or repeatedly alarming clients).

Stage II (Client-Side).
Each selected client $k \in N$, as a follower, decides:
(a) Participation. Whether or not to actually participate in FL, given the proposed payment $R$ and any potential penalty or overhead.
(b) Mini-Batch Size $B_k$. How many local data samples to use per training iteration, balanced against computation cost $s_k$ and potential reward.
(c) Privacy Noise Scale. If the client can adaptively choose $\varepsilon_k$ or the noise variance $\sigma_k$, it trades off between model accuracy and data privacy.
(d) Alarm-Triggering Strategy. Whether to raise alarms (and at what confidence level $\gamma_k$) based on local model testing.
or low-quality behaviors are discouraged.

These decisions collectively influence the final global model’s accuracy, robustness, and the net reward allocated to each participant.By jointly analyzing these utility functions in the two-stage game, CoSIFL allocates rewards and shapes client behaviors—balancing privacy, data quality, and security. This game-theoretic viewpoint ensures that even in adversarial conditions, well-intentioned clients receive fair compensation while malicious clients do not.

\section{Design of the CoSIFL algorithm}
\begin{figure}
	\centering
	\includegraphics[width=0.51\textwidth]{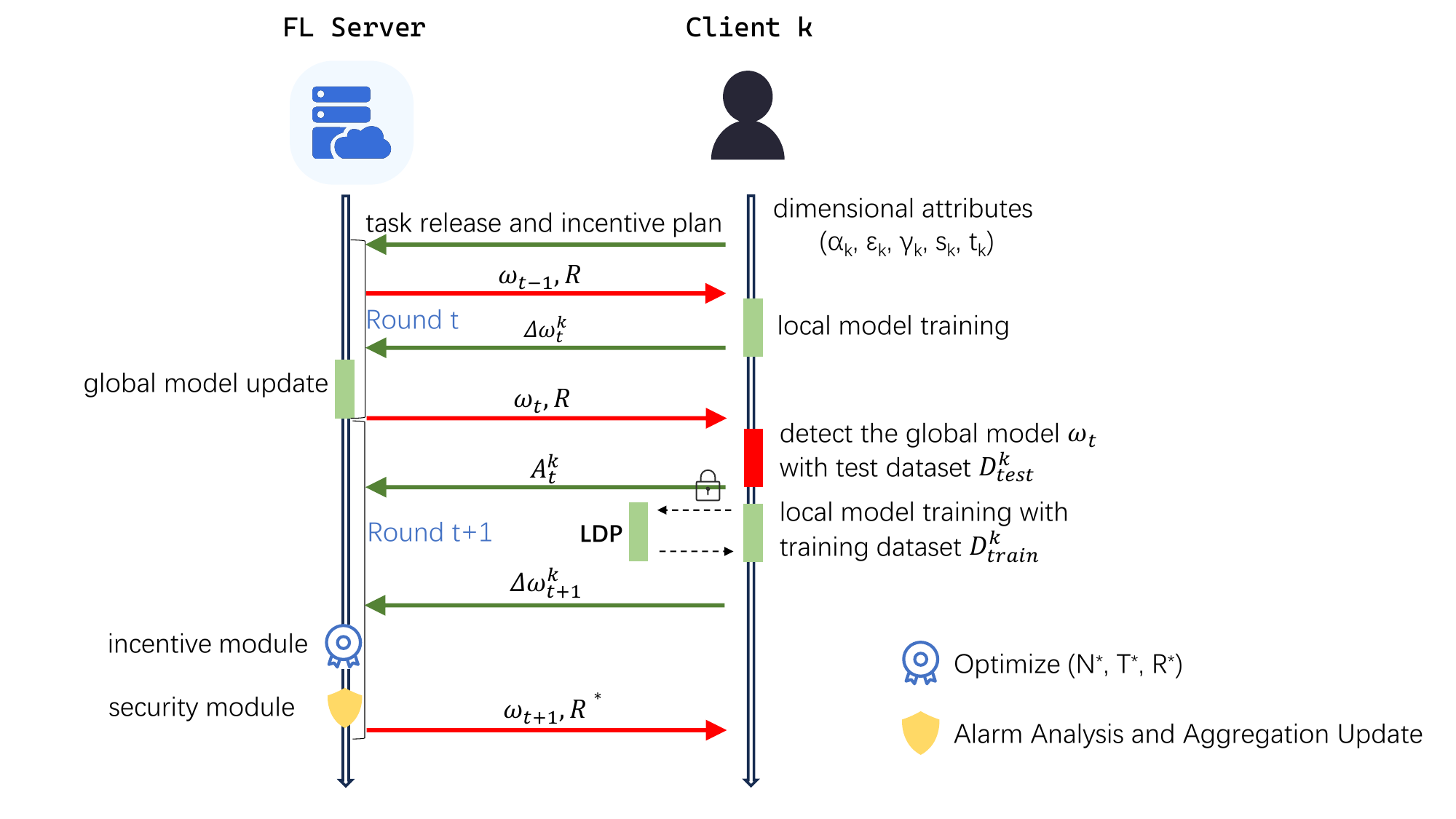}
	\caption{A schematic workflow of the proposed CoSIFL framework.}
	\label{CoSIFL's workflow}
\end{figure}

\subsection{Overview of CoSIFL}

Federated learning inherently faces not only security threats—such as Byzantine and inference attacks—but also struggles to motivate clients to actively participate when privacy concerns and resource consumption are at stake. Existing defensive methods~\cite{29,30} often concentrate on server-side detection and robust aggregation, limiting their effectiveness against both adversarial updates and data leakage, while neglecting how to incentivize clients to contribute high-quality data. To address these issues, we propose CoSIFL, a new comprehensive framework that integrates proactive security measures, LDP, and incentive mechanisms based on game theory.

Figure~\ref{CoSIFL's workflow} illustrates the structure of CoSIFL, which comprises both the server-end design and the client-end design. On the client side, two primary processes are deployed: a local training process (which can include LDP) and a proactive alarming process. Specifically, each client reserves a small portion of its local dataset as a private test set for anomaly detection. When the LDP mechanism is activated, the client injects properly tuned noise according to its privacy budget, mitigating the risk of inference attacks. Meanwhile, the alarming process examines each newly received global model against the client’s private test set and local model. If it detects anomalous behavior, it triggers an alarm to the server.

On the server side, CoSIFL coordinates the two-phase detection workflow and an incentive module. Upon receiving local updates and alarm signals, the server first applies an alarm-based filtering to remove obviously malicious updates, then performs Byzantine-robust aggregation on the remaining updates. Concurrently, the server runs a Stackelberg game-based incentive scheme to compensate clients. Drawing upon a Tullock-like contest~\cite{28} principle, CoSIFL rewards clients based on their data contribution, alarm accuracy, and privacy cost, encouraging them to adopt strategies (e.g., batch size selection, noise injection levels) beneficial to the overall system performance. By coupling proactive security and game-theoretic incentives, CoSIFL can effectively safeguard the global model while ensuring that privacy and resource constraints do not deter clients from participating.

\subsection{Client-Side Algorithm}

In the client part of Figure~\ref{CoSIFL's workflow}, each client verifies whether the newly received global model \(\omega_t\) is potentially poisoned, then decides batch size and LDP settings, and finally uploads both its alarm status \(A_k^t\) and local update \(\Delta \omega_k^t\) to the FL server in each communication round. In CoSIFL, each client also preserves a local test dataset and caches its previous local model. For simplicity, we use the client \(k\) to represent a generic participant, which could be either malicious or benign. Algorithm 1 presents the pseudo-code for the CoSIFL client-end alarming and training processes. A detailed description follows:

Step 1 (Lines 1–6): Checking the Global Model. 
Client \(k\) enters the Alarm() procedure, comparing the accuracy \(\omega_t^{\mathrm{acc}}\) of the global model on its local test set with the cached local model \(\omega_{t}^{k}\). Let \(\omega_t^{\mathrm{acc}}\) be the accuracy of the global model, and \(\omega_{t}^{k,\mathrm{acc}}\) be the cached local model’s accuracy. If 
$
\omega_t^{(\mathrm{acc})} 
\;\ge\; 
\omega_{t}^{k,\mathrm{acc}} \,\cdot\, (1 - C_c),
$
where \(C_c\) dictates how tolerant the client is to performance differences between the global model and its local model, then client \(k\) deems \(\omega_t\) trustworthy, sets \(A_k^t = 0\), and initializes the next-round local model with \(\omega_t\). Otherwise, if 
$
\omega_t^{\mathrm{acc}} < \omega_{t}^{k,\mathrm{acc}} \cdot (1 - C_c),
$
the client regards \(\omega_t\) as suspicious, triggers an alarm by setting \(A_k^t = 1\), and reverts to the previously cached local model \(\omega_{t}^{k}\) as the starting point for this round’s local training.
No matter what the alarm status is, the client sends \(A_k^t\) back to the server via a secure channel~\cite{31}, ensuring that the alarm information cannot be tampered with.

Step 2 (Lines 7–10): Client Game – Best Response Batch Size (Nash equilibrium context). 
In this step, each client \(k \in N\) chooses the batch size \(B_k\) in a Tullock contest-based utility setting, aiming to maximize its local gain:
1.Nash Equilibrium Objective: 
Given the total reward \(R\) announced by the server and other clients’ potential contributions \(\{B_j\}_{j\neq k}\), client \(k\) solves:
\begin{equation}
	\max_{B_k \ge 0} \quad U_k 
	= 
	\frac{\alpha_k \,\gamma_k \,B_k}{\sum_{j\in K}\alpha_j \,\gamma_j\,B_j}
	\,R \;-\; s_k\,B_k.
\end{equation}

2.Best Response Computation: Analytical Method:  
If a closed-form solution exists (e.g., setting the derivative \(\frac{\partial U_k}{\partial B_k}\) to zero and solving for \(B_k\)), the client obtains \(B_k^*\). If \(B_k^* \le 0\), it chooses not to participate.Iterative Method: In more complex scenarios, the client may use numerical methods (e.g., gradient ascent or binary search) while assuming other clients’ \(B_j\) are fixed, incrementally approximating its best response.

3. Participation Decision in Nash Equilibrium: 
When all clients compute their respective \(B_k^*\), those with \(B_k^*>0\) will participate, whereas \(B_k^*=0\) indicates opting out.  
Once no single client can unilaterally improve its utility, the system reaches a Nash Equilibrium.
By incorporating such a game-theoretic approach, clients make an optimal batch-size decision that balances rewards and costs, ultimately boosting global training performance.

Step 3 (Lines 11–15): Defending Against Inference Attacks. 
Since FL may face multiple types of attacks, CoSIFL optionally enables LDP to counter inference attacks.In CoSIFL, the local training includes:Gradient Calculation  
\begin{equation}
	\partial g^{k} = \nabla f\bigl(B_j^{k},\, g^{k}\bigr),
\end{equation}
where \(f\) is the local loss, \(g^{k}\) is the current model weight.

2.Gradient clipping is designed to limit the maximum norm of the gradients uploaded by each client, thereby reducing or controlling the risk of privacy leakage.
\begin{equation}
	\partial {{\mathop g\limits^-} ^k} = \frac{\partial g^{k}}{\max\bigl(1, \,\| \partial g^{k}\|_2/C^{k}\bigr)},
\end{equation}
with \(C^{k}\) as the clipping threshold.

3. Noise Injection  
\begin{equation}
	\partial {{\mathop g\limits^ \sim}  ^k} = \partial {{\mathop g\limits^-} ^k} + \mathcal{N}\!\bigl(0,\,(\sigma^{(k)})^2 (C^{(k)})^2 \mathbf{I}\bigr),
\end{equation}
where \(\sigma^{k}\) is the noise multiplier,${\sigma _k} = \frac{{2\sqrt {2\ln (1.25/\delta )} \eta CT}}{{{B_k}{\varepsilon _k}}}$ and $\mathbf{I}$ is identity matrix.

4. Model Update. 
The clipped, noised gradient $\partial {{\mathop g\limits^ \sim}  ^k}$ is used to update the local model \(g^{(k)}\). After iterating over all local batches, the client obtains the round-\(t\) model \(g_{t+1}^{(k)}\).

Step 4 (Lines 16–18): Upload Model Update and Return
Finally, client \(k\) computes \(\Delta \omega_k^{t+1} = \omega_k^{t+1} - \omega_t\) and uploads \(\Delta \omega_k^{t+1}\) to the server. The newly obtained local model \(\omega_k^{t+1}\) is cached for reference in the next round’s alarm process.
By integrating proactive alarming, local DP, and game-based batch-size selection, CoSIFL ensures robust and privacy-preserving federated training even in adversarial settings.

\begin{algorithm}[ht]
	\caption{CoSIFL Client-Side Procedure}
	\label{alg:CoSIFL-Client}
	\begin{algorithmic}[1]
		
		\Require 
		Global model $\omega_t$ from last round, cached local model $\omega_t^{k}$, 	local attributes $(\alpha_k,\varepsilon_k,\gamma_k, s_k, t_k)$, alarm threshold $C_c$, local data $D_k$, test set $D_k^{\mathrm{test}}$, total reward $R$
		\Ensure 
		Local update $\Delta \omega_k^{t+1}$, alarm status $A_k^t$
		
		\Statex \textbf{// Step 1: Alarm Trigger}
		\State $(\mathrm{\omega _t^{acc}}, \omega _t^{k,acc}) \gets (\text{AccTest}(\omega_t,D_k^{\mathrm{test}}), \text{AccTest}(\omega_t^{k},D_k^{\mathrm{test}}))$
		\If{$\omega _t^{acc} \ge \omega _t^{k,acc} \cdot (1 - C_c)$}
		\State $A_k^t \gets 0,\; g \gets \omega_t \;\;$ \Comment{g is the initialization model for training.}
		\Else
		\State $A_k^t \gets 1,\; g \gets \omega_t^{k} \;\;$ 
		\EndIf
		
		\Statex \textbf{// Step 2: Client Game -- Best Response Batch Size}
		\State $B_k \gets \text{ComputeBestResponse}(R,\alpha_k,\gamma_k,\varepsilon_k,s_k)$ 
		\If{$B_k = 0$}
		\State \textbf{return} (No participation)
		\EndIf
		
		\Statex \textbf{// Step 3: Local Training with or w/o LDP}
		\If{$\text{useLDP}$} \Comment{Defense against inference attacks}
		\State $g_{t}^{k} \gets \text{LocalTrainWithLDP}(g, D_k, B_k, \varepsilon_k)$
		\Else
		\State $g_{t}^{k} \gets \text{LocalTrain}(g, D_k, B_k)$
		\EndIf
		
		\Statex \textbf{// Step 4: Upload Update}
		\State $\Delta \omega_k^{t+1} \gets g_{t}^{k} - \omega_t$
		\State Send $\{ \Delta \omega_k^{t+1}, A_k^t \}$ to server 
		\State $\omega_{t+1}^{k} \gets g_{t}^{k}$ \Comment{Cache new local model for next round}
		
	\end{algorithmic}
\end{algorithm}

\subsection{Server-Side Algorithm}\label{Server-Side Algorithm}

\subsubsection{Approximate Incentive Optimization (Lines 1-2)}

The server possesses a pool of candidate clients \( K \), yet not all of them must be selected, as clients exhibit heterogeneity in computing capacity, network latency, privacy requirements, data quality, etc.
Given a limited training time \( D_{\mathrm{total}} \), the server aims to choose an appropriate subset \( N \subseteq K \) while specifying the global iteration count \( T \) and the total reward \( R \), in order to minimize the overall server cost \( C_{\text{server}}(N,\,T,\,R) \).
Exhaustively searching all combinations of (client subsets, iteration counts, reward allocations) is typically exponential in scale, hence a near-optimal heuristic is required to find a decent solution.Therefore, we used Pareto selection to select the subset of clients $N^*$ the global number of rounds $T^*$ and the total incentive $R^*$ that find the optimal, as detailed in Section~\ref{Approximate Incentive Optimization}.

\subsubsection{Alarm-Silence Analysis and Aggregation Update}

Due to the inherent vulnerabilities of FL, the server cannot fully trust any local model updates uploaded by clients or blindly accept their alarm signals. In the $(t+1)$-th communication round, the server conducts a two-phase detection process: it first verifies whether the previously aggregated global model \(\omega_t\) is poisoned, then checks whether the newly received model updates \(\{\Delta \omega_{t+1}^k, k \in N\}\) are poisoned. Before performing global model aggregation, the server must analyze the alarm information from clients, which falls into three cases:

Case 1 (Lines 11–13): 
If \(\forall k \in N,\, A_k^t = 0\) (no client raises an alarm), then \(\omega_t\) might be unpoisoned, but it is also possible \(\omega_t\) was compromised but not yet detected. Further verification would occur in the next round. With no alarms present, the server treats the global model \(\omega_t\) from round \(t\) as safe and uses it directly for updating.

When one or more clients raise alarms, the server evaluates the accuracy of the alarming clients’ local models (where \(k \in S_a\)), identifies the maximum accuracy among them, \(\max\{\omega_t^{k}, k \in S_a\}\), and then uses a threshold \(C_s\) to measure how each alarming client’s accuracy differs from that maximum.

Case 2 (Lines 14–21):  
If all alarming clients in \(S_a\) share a similar accuracy, i.e.,
\begin{equation}\label{The accuracy among Sa is similar}
	\forall k \in S_a,\;\max(\omega_t^{k,\mathrm{acc}}\text{ in }S_a)\cdot(1 - C_s)<\omega_t^{k,\mathrm{acc}},
\end{equation}
the server compares the accuracy of the alarming clients \(S_a\) with that of the silent clients \(S_s\), in order to determine which set is benign (\(S_b\)).
If \(\max(\omega_t^{k,\mathrm{acc}}\text{ in }S_s)\) is higher than or roughly equal to \(\max(\omega_t^{k,\mathrm{acc}}\text{ in }S_a)\cdot(1 - C_s)\), i.e.,
\begin{equation}\label{Sa’s alarms to be false}
	\max(\omega_{t}^{\mathrm{k,acc}} \text{ in } S_a)\cdot (1 - {C_s})
	\le \max(\omega_{t}^{\mathrm{k,acc}} \text{ in } S_s),
\end{equation}
the server considers \(S_a\)’s alarms to be false. The rationale is that alarming clients, who rely on their local models, believe the global model is unsafe; however, if the silent clients (who accept \(\omega_t\)) exhibit a higher accuracy, the alarms must be incorrect. The server then treats the silent clients as benign, deems \(\omega_t\) unpoisoned, and proceeds with \(\omega_t\) for the update (Lines 16–17).
Conversely, if
$
\max(\omega_t^{k,\mathrm{acc}}\text{ in }S_a)\cdot(1 - C_s)>\max(\omega_t^{k,\mathrm{acc}}\text{ in }S_s),
$
all silent clients’ updates are deemed “bad,” the alarming clients are benign, and they believe \(\omega_t\) has been poisoned. As a result, the server reverts to the previous global model \(\omega_{t-1}\) for aggregation (Lines 19–20).

Case 3 (Lines 22–25):  
If there exists \(k\in S_a\) such that
$
\max(\omega_t^{k,\mathrm{acc}}\text{ in }S_a)\cdot(1 - C_s)\,\ge\,\omega_t^{k,\mathrm{acc}},
$
the accuracy among alarming clients diverges significantly, meaning \(S_a\) could contain both benign and malicious clients. The server thus deems \(\omega_t\) untrustworthy. If an alarming client’s accuracy is close to the maximum in \(S_a\), i.e., 
$
\max(\omega_t^{k,\mathrm{acc}}\text{ in }S_a)\cdot(1 - C_s)<\omega_t^{k,\mathrm{acc}},
$
that client is considered benign. Because benign clients will always alarm upon detecting poisoning, the server assumes no benign clients exist among the silent ones in this case, it ignores all silent clients. Finally, it reverts to \(\omega_{t-1}\) for model aggregation.

Penalty and Reward Mechanisms  
To reduce needless client computation overhead and strengthen defenses, the server enforces both penalty and reward mechanisms:
Penalty Mechanism: Malicious clients are penalized and their penalty count is recorded. Once a client’s penalty count exceeds \(C_p\), the server designates it as malicious by default—banning it from further training (and removing it from \(N^*\) if needed).
Reward Mechanism: Banned clients have a certain chance to rejoin. If, in some communication round, a previously banned client is judged benign by the server, its penalty count is reduced; if its penalty count then falls below \(C_p\), the client is once again allowed to participate in training.

\begin{algorithm}[ht]
	\caption{CoSIFL: Server-Side Procedure}
	\label{alg:CoSIFL-Server}
	\begin{algorithmic}[1]
		
		\Require  Candidate client set $K$, max training rounds $t_{\max}$, time budget $D_{\mathrm{total}}$, security thresholds $C_s$
		
		\Ensure 
		Final global model $\omega_{T}$, Chosen subset $N^*,$ global iteration $T^*$, total reward $R^*$
		
		\Function{ServerTraining}{$K,t_{\max},D_{\mathrm{total}}$}
		\State $ N^*, T^*,\, R^* \gets \text{ApproxClientSelection}(K,t_{\max},D_{\mathrm{total}})$ 
		\For{$t=1$ to $T^*$}
		\State Broadcast $\omega_{t}, R^*$ to clients in $N^*$
		\State Collect $\{A_k^t\}, \{\Delta\omega_k^{t+1}\}$ from each $k{\in}N^*$ by Alg.\ref{alg:CoSIFL-Client}
		\State $\omega_{t+1} \gets \text{ServerDetect}\bigl(\omega_{t-1},\omega_{t}, \{A_k^t\}, \{\Delta\omega_k^{t+1}\}\bigr)$
		\EndFor
		\State \Return $\omega_{T^*}$ 
		\EndFunction
		
		\vspace{2mm}
		
		\Function{ServerDetect}{$\bigl(\omega_{t-1}, \omega_{t}, \{A_k^t\}, \{\Delta\omega_k^{t+1}\}\bigr)$}
		\If{all $A_k^t=0$}\Comment{No alarms}
		\State \(\omega_{t+1} \gets \omega_{t} + \sum_{k\in N^*}\Delta \omega_k^{t+1}\) 
		\State \Return $\omega_t$
		\ElsIf{Satisfy Formula~\ref{The accuracy among Sa is similar} }
		\If{Satisfy Formula~\ref{Sa’s alarms to be false}}
		
		\State Treat $S_s$ as benign clients: ${S_b} \leftarrow {S_s}$
		\State ${\omega _{t + 1}} \leftarrow {\omega _t} + \sum\nolimits_{k \in {S_b}} {{\eta _k}} \Delta \omega _{t + 1}^k$
		\Else
		\State Treat $S_a$ as benign clients: ${S_b} \leftarrow {S_a}$
		\State ${\omega _{t + 1}} \leftarrow {\omega _{t-1}} + \sum\nolimits_{k \in {S_b}} {{\eta _k}} \Delta \omega _{t + 1}^k$
		
		\EndIf
		\Else \Comment \text{The accuracy among $S_a$ is not similar.}
		\State Treat $S_a$ as benign clients: ${S_b} \leftarrow {S_a}$
		\State ${\omega _{t + 1}} \leftarrow {\omega _{t-1}} + \sum\nolimits_{k \in {S_b}} {{\eta _k}} \Delta \omega _{t + 1}^k$
		\EndIf
		\State \Return $\omega_{t+1}$
		\EndFunction
		
	\end{algorithmic}
\end{algorithm}

\section{Approximate Incentive Optimization}\label{Approximate Incentive Optimization}

Based on equation~\ref{server's cost}, when considering the Nash equilibrium of the clients’ game, the server needs to solve the following problem:
\begin{equation}\label{min_server_cost}
	\min_{N,T,R} C_{\text{server}}, 
	\quad 
	\text{s.t. } N \subseteq K, \ 
	T \in \left[1, \left\lfloor \frac{D_{\text{total}}}{t_{\max}^{N^*}} \right\rfloor \right].
\end{equation}
To achieve the Stackelberg equilibrium, we seek an optimal solution that minimizes the server’s cost (the server’s best response to the clients’ equilibrium). However, finding the optimal solution to problem (\ref{min_server_cost}) is very challenging for the following reasons:

1. The client selection problem (choosing \(N\) from \(K\)) is coupled with the total number of global iterations \(T\), and \(T\) depends on each client’s response time \(t_k\) as well as client attributes (non-IID degree, privacy noise, alarm reliability, data usage cost).
2. Even after determining the chosen client set \(N\), it is still non-trivial to figure out \((T, R)\), mainly because of the complex form of \(C_{\text{server}}\) in equation~\ref{server's cost}.

Hence, pursuing a Stackelberg equilibrium requires extensive computation, which is undesirable for the server. To overcome these difficulties, we propose an approximately optimal method to solve (\ref{min_server_cost}). The core idea is: first propose an efficient client selection criterion; then, for the selected client set, derive the optimal strategy \((T, R)\). In practice, since \(T\) must lie in a finite integer set, we can fix \(T\) to determine the optimal \(R\), and then choose the \(T\) (among all possible values) that yields the minimum server cost.

Given a fixed client set \(N\), the optimal global iteration count \(T\) and total reward \(R\).	
First, suppose the chosen client set \(N\) is known. We aim to derive the optimal \((T, R)\) under this fixed \(N\). Due to the complexity of \(C_{\text{server}}\), it is difficult to obtain a closed-form solution directly, so we instead derive the following implicit solution.

\begin{proposition}~\cite{15}
	Given the chosen client set \(N\) and a fixed number of iterations \(T\), there is a unique equilibrium participating client set \(N^*\). When \(R > 0\), \(C_{\text{server}}\) is a convex function of \(R\). By setting the first-order derivative \(\frac{\partial C_{\text{server}}}{\partial R} = 0\), the optimal \(R^*\) is obtained by solving the root of the following equation:
	\begin{equation}\label{R's optimal}
		\kappa_4 R^3 + \kappa_5 R + \kappa_6 = 0,
	\end{equation}
	
	where
	$
	\kappa_4 
	= \gamma_5 \, T 
	\sum_{k \in N^*} \alpha_k \nu_k 
	\Bigl[
	1 - \frac{L_k}{\,(|N^*| - 1)\sum_{k \in N^*} L_k}
	\Bigr],
	\kappa_5 
	= -\gamma \gamma_3 
	(1 - \phi^T) 
	\frac{|N^*|}{Y(N^*)},
	\kappa_6 
	= -\gamma \gamma_2 
	(1 - \phi^T) 
	\sum_{k \in N^*} \frac{2}{\,T^2 [Y(N^*)]^2} \zeta_k^2.$
	The cubic equation~\ref{R's optimal} can be solved via Cardano’s formula~\cite{39}. Then, for a given \(N\) (and \(N^*\)), since \(T\) is an integer, we can search over 
	$
	T = 1, 2, \ldots, \left\lfloor \frac{D_{\text{total}}}{t_{\max}^{N^*}} \right\rfloor
	$
	to find the value of \(T\) that minimizes the server’s cost.
\end{proposition}

\textbf{Approximately Optimal Client Selection}

It is not straightforward to directly determine the optimal client selection strategy, because the server needs to account for various client attributes (non-IID degree, privacy noise, alarm reliability, data usage cost and response times), and the nature of the Nash equilibrium itself. Below, we propose an approximately optimal client selection scheme.
From (\ref{server's cost}), we observe that a fully trained model requires a large number of global iterations \(T\), at which point the negative impact of the privacy term becomes dominant. A large total batch size can simultaneously mitigate the errors caused by data sampling and privacy noise. Therefore, our key idea is: under the constraint that the response time of each chosen client does not exceed a given \(t_{\max}\), we strive to maximize the total batch size after client competition.

Based on the nature of the Nash equilibrium in the client game, if the total reward \(R\) is fixed, then a larger conversion rate \(Y(N^*)\) implies a larger total batch size \(B(N^*)\). Formally, we have the following conclusion:

\begin{theorem}\cite{15}\label{theorem 1}
	Suppose we have a chosen client set \(N\) and another set \(\widetilde{N}\) where \(\widetilde{N} \subseteq N\). Let their respective Nash-equilibrium participating client sets be \(N^*\) and \(\widetilde{N}^*\). Then:
	$
	Y(\widetilde{N}^*) \;\le\; Y(N^*).
	$ 
\end{theorem}

\begin{figure}
	\centering
	\includegraphics[width=0.35\textwidth]{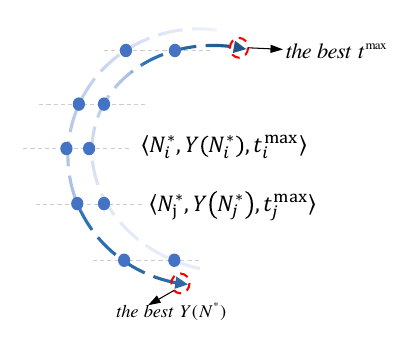}
	\caption{Pareto selection.}
	\label{Pareto selection}
\end{figure}

Theorem~\ref{theorem 1} indicates that, as long as the time constraints permit, expanding an already chosen client set is beneficial for improving the total batch size under the equilibrium of the client game. At the same time, selecting more clients could cause the maximum response time to increase. As shown in Figure~\ref{Pareto selection}, a higher conversion rate corresponds to a longer response time, and a lower response time corresponds to a lower conversion rate. Based on these two conflicting objectives, we use the concept of Pareto optimality to address the trade-off.

\begin{definition}
	Suppose for two chosen client sets \(N\) and \(\widetilde{N}\), their respective Nash-equilibrium participating client sets are \(N^*\) and \(\widetilde{N}^*\). If
	$
	Y(\widetilde{N}^*) \;\ge\; Y(N^*), 
	\quad 
	t_{\max}^{\widetilde{N}^*} \;\le\; t_{\max}^{N^*},
	$
	and at least one of the following holds strictly,
	$
	Y(\widetilde{N}^*) \;>\; Y(N^*) 
	\quad \text{or} \quad 
	t_{\max}^{\widetilde{N}^*} \;<\; t_{\max}^{N^*},
	$
	we say \(\widetilde{N}^*\) Pareto-dominates \(N^*\). If there is no \(\widetilde{N}^* \subseteq K\) that can Pareto-dominate \(N^*\), then \(N^*\) is called a Pareto selection.
\end{definition}

We denote by \(\mathcal{H}\) the set of all Pareto selections. Below is a summary of the approximate optimal strategy procedure for minimizing the server’s cost:

1. Initialization: Set \(\mathcal{H} = \varnothing\).  
First, select all candidate clients \(K\) as the chosen set and compute its Nash equilibrium, yielding the participating client set \(N_i^*\). Let the corresponding maximum response time be \(t_{\max}^i = \max_{k \in N_i^*} (t_k)\).  
By Theorem~\ref{theorem 1}, for any subset \(N \subseteq K\), the set \(N_i^*\) achieves the largest conversion rate \(Y(N_i^*)\). Clearly, \(N_i^*\) must be one of the Pareto selections.  
Next, remove from \(K\) any clients whose response time exceeds \(t_{\max}^i\), i.e., update 
$
K \leftarrow K \setminus \{\, a \mid t_a > t_{\max}^i \}, 
$
because if there exists a client \(a\) with \(t_a > t_{\max}^i\) and \(a \in \widetilde{N}\), then \(N_a^*\) will dominate \(\widetilde{N}\).  
Record the Pareto selection \(N_i^*\) along with its related information \(\langle N_i^*, Y(N_i^*), t_{\max}^i \rangle\), and update 
$
\mathcal{H} \leftarrow \mathcal{H} \cup \{\langle N_i^*, Y(N_i^*), t_{\max}^i\rangle\}.
$

2.Search for the Next Pareto Selection:
We continue searching by “relaxing” the maximum response time. Specifically, remove from \(K\) the client \(a\) with the largest response time in \(N_i^*\): 
$
a = \arg\max_{k \in N_i^*} t_k,  
K \leftarrow K \setminus \{a\}.
$  
Using the new set \(K\), compute the new Nash equilibrium to obtain another participating client set \(N_j^*\), its conversion rate \(Y(N_j^*)\), and the maximum response time \(t_{\max}^j = \max_{k \in N_j^*}(t_k)\).  
Insert \(\langle N_j^*, Y(N_j^*), t_{\max}^j\rangle\) into \(\mathcal{H}\).  
Repeat the above steps until \(|K| < 2\). By Theorem~\ref{theorem 1}, at each step, for a given maximum response time \(t_{\max}\), the set obtained always yields the largest \(Y\); hence each selection is inevitably a Pareto selection (see Figure~\ref{Pareto selection}).

3. Compare All Pareto Selections:  
For each Pareto selection in \(\mathcal{H}\) (i.e., each fixed set \(N\)), compute the optimal \((T, R)\) that minimizes the server’s cost.  
Finally, compare the minimum server costs achieved by all Pareto selections and choose the optimal one.
Select the configuration \(\{N^*,\,T^*,\,R^*\}\) yielding the smallest \(C_{\text{server}}\) as the near-optimal result.

\section{Stability of the Global Model in CoSIFL}

The stability and convergence of the global model in CoSIFL are critical to ensuring that the FL system achieves reliable performance despite challenges such as Byzantine attacks, inference threats, and DP noise. In this section, we provide a rigorous theoretical analysis of the global model’s stability, incorporating the effects of proactive alarming, Byzantine-robust aggregation, and local DP. We derive convergence bounds and stability conditions using established assumptions and mathematical tools from optimization theory and stochastic processes.

\subsection{\textbf{Assumptions}}

To facilitate the analysis, we adopt the following standard assumptions commonly used in FL convergence studies:

1. Bounded Gradients: For each client \( k \), the gradient of the local loss function \( \nabla F_k(\omega) \) is \( L \)-Lipschitz continuous, i.e., 
\[
\|\nabla F_k(\omega) - \nabla F_k(\omega')\| \leq L \|\omega - \omega'\|, \forall \omega, \omega'.
\]
Additionally, the gradient norm is bounded: \( \|\nabla F_k(\omega)\| \leq G \).

2. Non-IID Data: The local loss \( F_k(\omega) \) deviates from the global loss \( F(\omega) = \sum_{k \in K} p_k F_k(\omega) \) due to data heterogeneity, with a bounded divergence: 
$
\|\nabla F_k(\omega) - \nabla F(\omega)\| \leq \lambda_k,
$
where \( \lambda_k \) quantifies the non-IID degree of client \( k \).

3. DP Noise: Each client \( k \) adds Gaussian noise \( \mathcal{N}(0, \sigma_k^2 I) \) to its gradients to satisfy \( (\varepsilon_k, \delta) \)-DP, where \( \sigma_k = \frac{2\sqrt{2 \ln (1.25/\delta)} \eta C T}{B_k \varepsilon_k} \), \( C \) is the clipping threshold, \( T \) is the number of global iterations, and \( B_k \) is the batch size.

4. Malicious Clients: A fraction \( f < 0.5 \) of clients may be malicious, submitting arbitrary updates. The Byzantine-robust aggregation ensures that their impact is mitigated.

5. Strong Convexity: The global loss \( F(\omega) \) is \( \mu \)-strongly convex, i.e., 
$
F(\omega') \geq F(\omega) + \langle \nabla F(\omega), \omega' - \omega \rangle + \frac{\mu}{2} \|\omega' - \omega\|^2.
$

\subsection{\textbf{Global Model Update}}

In CoSIFL, the global model \( \omega_t \) at round \( t \) is updated as:
$
\omega_{t+1} = \omega_t + \eta_t \Delta \omega_t,
$
where \( \eta_t \) is the learning rate, and \( \Delta \omega_t \) is the aggregated update from benign clients \( N_b \subseteq N^* \):
$
\Delta \omega_t = \frac{1}{|N_b|} \sum_{k \in N_b} \Delta \omega_k^t.
$
Each client \( k \)’s local update is:
$
\Delta \omega_k^t = -\eta_k \left( \nabla F_k(\omega_t) + \xi_k^t \right),
$
where \( \eta_k \) is the local learning rate, and \( \xi_k^t \sim \mathcal{N}(0, \sigma_k^2 I) \) is the DP noise. The set \( N_b \) is determined by the proactive alarming and alarm-silence analysis, excluding malicious updates.

\subsection{\textbf{Convergence Analysis}}

We aim to bound the expected deviation of \( \omega_t \) from the optimal model \( \omega^* = \arg\min F(\omega) \) and prove convergence. Define the expected loss gap as:
$
\mathbb{E}[F(\omega_t)] - F(\omega^*).
$

Step 1: Local Update Impact.
For a benign client \( k \in N_b \), the local update with DP noise is:
$
\Delta \omega_k^t = -\eta_k \nabla F_k(\omega_t) - \eta_k \xi_k^t.
$
The expected local gradient is unbiased due to the zero-mean noise: \( \mathbb{E}[\xi_k^t] = 0 \). However, the noise introduces variance:
$
\mathbb{E}[\|\xi_k^t\|^2] = d \sigma_k^2,
$
where \( d \) is the model dimension.

Step 2: Global Update Decomposition.
Substitute the local updates into the global aggregation:
$
\Delta \omega_t = -\frac{\eta_k}{|N_b|}  \times  \\
\sum_{k \in N_b} \nabla F_k(\omega_t) 
- \frac{\eta_k}{|N_b|} \sum_{k \in N_b} \xi_k^t.
$
Define the aggregated benign gradient as:
$
\overline{\nabla F}(\omega_t) = \frac{1}{|N_b|} \sum_{k \in N_b} \nabla F_k(\omega_t).
$
Thus:
$
\Delta \omega_t = -\eta_k \overline{\nabla F}(\omega_t) - \eta_k \overline{\xi_t},
$
where \( \overline{\xi_t} = \frac{1}{|N_b|} \sum_{k \in N_b} \xi_k^t \), and:
$
\mathbb{E}[\overline{\xi_t}] = 0, \quad \mathbb{E}[\|\overline{\xi_t}\|^2] = \frac{1}{|N_b|^2} \sum_{k \in N_b} d \sigma_k^2.
$

Step 3: Distance to Optimum.
Consider the distance \( \|\omega_{t+1} - \omega^*\|^2 \):
$
\omega_{t+1} = \omega_t + \eta_t \Delta \omega_t = \omega_t - \eta_t \eta_k \overline{\nabla F}(\omega_t) - \eta_t \eta_k \overline{\xi_t}.
$
Then:
$
\|\omega_{t+1} - \omega^*\|^2 = \|\omega_t - \omega^* - \eta_t \eta_k \overline{\nabla F}(\omega_t) - \eta_t \eta_k \overline{\xi_t}\|^2.
$
Expanding the norm and taking expectations:
$
\mathbb{E}[\|\omega_{t+1} - \omega^*\|^2] = \|\omega_t - \omega^*\|^2 - 2 \eta_t \eta_k \langle \omega_t - \omega^*, \overline{\nabla F}(\omega_t) \rangle + \eta_t^2 \eta_k^2 \mathbb{E}[\|\overline{\nabla F}(\omega_t) + \overline{\xi_t}\|^2].
$
Since \( \mathbb{E}[\overline{\xi_t}] = 0 \):
$
\mathbb{E}[\|\overline{\nabla F}(\omega_t) + \overline{\xi_t}\|^2] = \|\overline{\nabla F}(\omega_t)\|^2 + \mathbb{E}[\|\overline{\xi_t}\|^2].
$

Step 4: Strong Convexity and Gradient Bounds.
Using the strong convexity of \( F(\omega) \):
$
\langle \nabla F(\omega_t), \omega_t - \omega^* \rangle \geq F(\omega_t) - F(\omega^*) + \frac{\mu}{2} \|\omega_t - \omega^*\|^2.
$
Due to non-IID data, relate \( \overline{\nabla F}(\omega_t) \) to \( \nabla F(\omega_t) \):
$
\overline{\nabla F}(\omega_t) = \nabla F(\omega_t) + \frac{1}{|N_b|} \sum_{k \in N_b} (\nabla F_k(\omega_t) - \nabla F(\omega_t)).
$
Define the heterogeneity term:
$
\Delta_t = \frac{1}{|N_b|} \sum_{k \in N_b} \\
(\nabla F_k(\omega_t) - \nabla F(\omega_t)), \quad \|\Delta_t\| \leq \overline{\lambda} = \frac{1}{|N_b|} \sum_{k \in N_b} \lambda_k.
$
Thus:
$
\langle \omega_t - \omega^*, \overline{\nabla F}(\omega_t) \rangle = \langle \omega_t - \omega^*, \nabla F(\omega_t) \rangle + \langle \omega_t - \omega^*, \Delta_t \rangle.
$
Bound the heterogeneity term:
$
\langle \omega_t - \omega^*, \Delta_t \rangle \geq -\|\omega_t - \omega^*\| \|\Delta_t\| \geq -\|\omega_t - \omega^*\| \overline{\lambda}.
$

Step 5: Convergence Bound.
Substitute into the distance expression:
$
\mathbb{E}\!\left[\|\omega_{t+1} - \omega^*\|^2\right] \le
\|\omega_t - \omega^*\|^2
- 2 \eta_t \eta_k \Big( F(\omega_t) - F(\omega^*)
+ \tfrac{\mu}{2}\|\omega_t - \omega^*\|^2 
- \|\omega_t - \omega^*\|\,\overline{\lambda} \Big)
+ \eta_t^2 \eta_k^2 \Big( \|\overline{\nabla F}(\omega_t)\|^2
+ \tfrac{d}{|N_b|^2} \sum_{k \in N_b} \sigma_k^2 \Big).
$
Since \( \|\overline{\nabla F}(\omega_t)\| \leq G \), and assuming \( \eta_t \eta_k < 1/L \) to ensure descent:
$
\mathbb{E}[\|\omega_{t+1} - \omega^*\|^2] \leq (1 - \eta_t \eta_k \mu) \|\omega_t - \omega^*\|^2 - 2 \eta_t \eta_k (F(\omega_t) - F(\omega^*)) + 2 \eta_t \eta_k \overline{\lambda} \|\omega_t - \omega^*\| + \eta_t^2 \eta_k^2 \left( G^2 \\
+ \frac{d}{|N_b|^2} \sum_{k \in N_b} \sigma_k^2 \right).
$
Define \( r_t = \mathbb{E}[\|\omega_t - \omega^*\|^2] \). Then:
$
r_{t+1} \leq (1 - \eta_t \eta_k \mu) r_t + 2 \eta_t \eta_k \overline{\lambda} \sqrt{r_t} + \eta_t^2 \eta_k^2 \left( G^2 + \frac{d}{|N_b|^2} \sum_{k \in N_b} \sigma_k^2 \right).
$

Step 6: Stability and Convergence Rate.
For a constant learning rate \( \eta_t = \eta = \frac{1}{\mu T} \), \( \eta_k = 1 \), and \( T \) iterations:
$
r_{t+1} \leq \left(1 - \frac{1}{T}\right) r_t + \frac{2 \overline{\lambda}}{\mu T} \sqrt{r_t} + \frac{1}{\mu^2 T^2} \left( G^2 + \frac{d}{|N_b|^2} \sum_{k \in N_b} \sigma_k^2 \right).
$
Iterating this recurrence, and assuming \( r_0 = \|\omega_0 - \omega^*\|^2 \):
$
r_T \leq \left(1 - \tfrac{1}{T}\right)^T r_0 +
\sum_{t=0}^{T-1} \left(1 - \tfrac{1}{T}\right)^t
\Bigg( \tfrac{2 \overline{\lambda}}{\mu T} \sqrt{r_t} \\
+ \tfrac{G^2 + \tfrac{d}{|N_b|^2} \sum_{k \in N_b} \sigma_k^2}{\mu^2 T^2} \Bigg).
$
Since \( (1 - \frac{1}{T})^T \approx e^{-1} \), and bounding \( \sqrt{r_t} \leq \sqrt{r_0} \):
$
r_T \leq e^{-1} r_0 + \frac{2 \overline{\lambda} \sqrt{r_0}}{\mu} + \frac{G^2 + \frac{d}{|N_b|^2} \sum_{k \in N_b} \sigma_k^2}{\mu^2 T}.
$
The loss gap follows:
$
\mathbb{E}[F(\omega_T) - F(\omega^*)] \leq \frac{\mu}{2} r_T \leq \frac{\mu e^{-1}}{2} r_0 + \overline{\lambda} \sqrt{r_0} + \frac{G^2 + \frac{d}{|N_b|^2} \sum_{k \in N_b} \sigma_k^2}{2 \mu T}.
$

\textbf{Interpretation}
1. Initial Distance: The term \( \frac{\mu e^{-1}}{2} r_0 \) decays exponentially with \( T \), reflecting convergence from the initial model.
2. Heterogeneity: \( \overline{\lambda} \sqrt{r_0} \) captures the impact of non-IID data, mitigated by selecting clients with lower \( \lambda_k \).
3. Noise Impact: \( \frac{d}{|N_b|^2} \sum_{k \in N_b} \sigma_k^2 / (2 \mu T) \) shows that DP noise variance decreases with more benign clients and iterations, but increases with stricter privacy (smaller \( \varepsilon_k \)).

Robustness against malicious clients.
The proactive alarming and alarm-silence analysis ensure \( |N_b| \geq (1 - f) |N^*| \). If \( f < 0.5 \), the majority of updates are benign, and the aggregation excludes malicious contributions, maintaining stability. The penalty mechanism further reduces \( f \) over time by banning repeat offenders.
The global model in CoSIFL converges to a neighborhood of \( \omega^* \), with the error bounded by data heterogeneity and DP noise. The stability is guaranteed by Byzantine-robust aggregation and proactive alarming, ensuring resilience against attacks while balancing privacy and utility.

\section{Experiment}
\subsection{Experimental Settings}
\noindent \underline{\textbf{Hardware}}
A Ubuntu (20.04.1) machine equipped with 8 NVIDIA GeForce RTX 2080 Ti GPUs.

\noindent \underline{\textbf{Datasets/Models}}
We use Fashion-MNIST and CIFAR-10 to evaluate the experiments. Fashion-MNIST consists of 60,000 grayscale training images and 10,000 test images, covering 10 categories. CIFAR-10 contains 50,000 color training images and 10,000 test images. We use a CNN model for training on Fashion-MNIST and ResNet-18 for training on CIFAR-10.

\noindent \underline{\textbf{Evaluation Metrics}}

Model Performance: Accuracy is used as the evaluation metric.

Misclassification Confidence~\cite{40}: The effectiveness of defending against attacks under targeted attacks. The lower the value, the fewer the misclassifications, indicating better defense.

Area Under the Curve (AUC): AUC is used to measure the classification model's ability to distinguish between positive and negative samples.AUC = 1.0: The attacker correctly classifies almost all instances, indicating severe model leakage. AUC = 0.5: The attacker has the same performance as random guessing, meaning the model is well-protected.

\noindent \underline{\textbf{IID and Non-IID Data}}
IID Data: The training data is randomly split into $K$ shards, and these data shards are directly allocated to $K$ clients.
Non-IID Data: We follow the pattern in the adaptive attack literature~\cite{37} to construct non-IID data using the non-IID degree \( p \). Training data with label \( l \) is assigned to the \( l \)-th group of clients with probability \( p \). The higher the \( p \), the higher the degree of non-IID.

\noindent \underline{\textbf{Data Quality}}

High Data Quality: malicious proportion = 0.1, iid.
Medium Data Quality: malicious proportion = 0.4,non-iid = 0.4.
Low Data Quality: malicious proportion = 0.8,non-iid = 0.8.

\noindent \underline{\textbf{Discrimination Rules}}

ND (No Discrimination): Does not differentiate the data quality of clients (\( \alpha_k = 1, \nu_k = 1 \), the data is iid and the noise is zero), considers the client response time limit (chooses clients based on latency), and selects clients based on unit data usage cost.

NDT (No Discrimination and Training Time): Does not differentiate the data quality of clients, does not consider response time limits, and always selects all candidate clients, using a fixed global iteration count \( T \).

TPUC (Two Part Uniform Contract): TPUC~\cite{38} is a FL incentive mechanism based on contract theory. It divides clients into several types and provides two uniform contract options (paid and unpaid) for each type, allowing for client selection and reward distribution under conditions of information asymmetry.

\noindent \underline{\textbf{Comparison of defense methods against Byzantine attacks}}

Siren+~\cite{13}: Siren+ combines an active client alert mechanism in FL with LDP techniques, defending against both Byzantine attacks and inference attacks.

Krum/Multi-Krum~\cite{32}: In each communication round, the server computes the sum of Euclidean distances between each client’s model update and those of all other clients, and assigns each update a score (its total distance). It then selects the update(s) with the lowest score (i.e., closest to the other client updates) as trusted updates for aggregation, discarding the rest. Krum: selects only one best client. Multi-Krum: selects multiple clients until the remaining number is less than twice the number of malicious clients.

Coomed~\cite{33}: Instead of relying on the overall weight differences, the server takes the median of all client updates for each coordinate (each parameter dimension) to form new global model weights, thereby mitigating the negative impact of individual malicious updates.

FLTrust~\cite{26}: The server maintains a trusted “small root dataset” to train an auxiliary model, which is used to evaluate the reliability of the model updates submitted by clients. The server calculates the cosine similarity between each client’s weight update and the server-side auxiliary model, assigns a “trust score,” and then aggregates client updates based on this score.

FedAvg~\cite{01}: No defense measures applied.

FL baseline: Represents the normal performance of FL in the ideal case where all participating clients are honest and there are no attacks. It is used to verify the effectiveness of the defense methods.

\noindent \underline{\textbf{Inference attack methods}}

In the experiment, we tested three membership inference attack (MIA) methods.

MLP(Multi-Layer Perceptron Attack) MIA: Uses a neural network to learn complex patterns from model outputs to infer whether a sample was in the training set.  

Threshold(Threshold-Based Attack) MIA: Infers membership by applying a simple threshold on confidence scores or loss values.  

LR(Logistic Regression Attack) MIA: Applies logistic regression to classify model output features and identify member samples.

\subsection{Effectiveness in defending against untargeted attacks}
\begin{figure*}[htbp]
	\centering
	\subfigure[sign flipping attack]{
		\centering
		\begin{minipage}[b]{0.31\textwidth}
			\includegraphics[width=1\textwidth, trim={0cm 0cm 0cm 0cm}, clip]{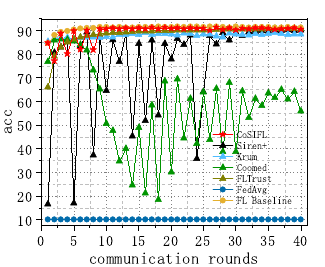}
		\end{minipage}
	}
	\subfigure[label flipping attack]{
		\centering
		\begin{minipage}[b]{0.31\textwidth}
			\includegraphics[width=1\textwidth, trim={0cm 0cm 0cm 0cm}, clip]{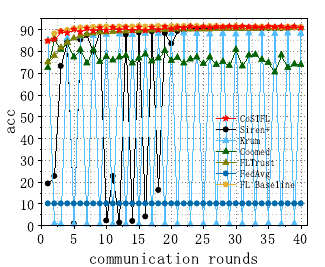}
		\end{minipage}
	}
	\subfigure[adaptive attack]{
		\centering
		\begin{minipage}[b]{0.31\textwidth}
			\includegraphics[width=1\textwidth, trim={0cm 0cm 0cm 0cm}, clip]{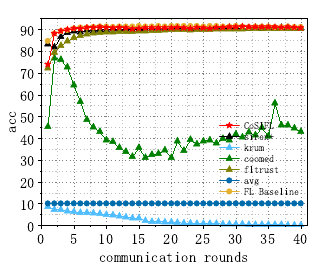}
		\end{minipage}
	}
	\caption{Untargeted attacks, K=10, high data quality}
	\label{Untargeted attacks}
\end{figure*}

We evaluated CoSIFL's defense against untargeted attacks (sign-flipping, label-flipping, adaptive attacks) under high, medium, and low data quality settings (\(K=10\) and \(K=50\) clients), comparing it with Siren+, Krum, Coomed, FLTrust, FedAvg, and the FL Baseline. Results are shown in Figures~\ref{Untargeted attacks},\ref{Untargeted attacks, non-iid=0.4},\ref{Untargeted attacks, non-iid=0.8}. Under high and medium data quality, Siren+ and FLTrust exhibit good defense performance, and even under low data quality, they can effectively mitigate attacks. This is because Siren+ also incorporates proactive alarming but lacks an incentive mechanism to encourage high-quality data contributions, leading to inferior performance compared to CoSIFL in low data quality scenarios. FLTrust relies on a server-side auxiliary model and trust scoring, performing well but slightly worse than CoSIFL under low data quality due to the auxiliary model's potential inability to accurately evaluate updates. Krum and Coomed completely lose their defense capabilities under low data quality. CoSIFL consistently outperforms all baseline methods across all settings, demonstrating strong defense even under low data quality, thanks to its proactive alarming, incentive mechanism, and robust aggregation, making it highly effective against untargeted attacks in adversarial federated learning environments.

\begin{figure*}[htbp]
	\centering
	\subfigure[sign flipping attack]{
		\centering
		\begin{minipage}[b]{0.31\textwidth}
			\includegraphics[width=1\textwidth, trim={0cm 0cm 0cm 0cm}, clip]{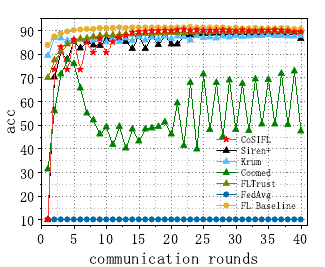}
		\end{minipage}
	}
	\subfigure[label flipping attack]{
		\centering
		\begin{minipage}[b]{0.31\textwidth}
			\includegraphics[width=1\textwidth, trim={0cm 0cm 0cm 0cm}, clip]{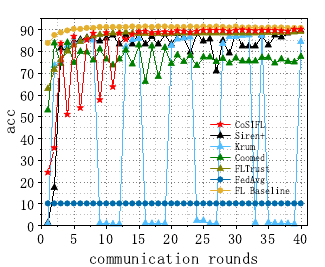}
		\end{minipage}
	}
	\subfigure[adaptive attack]{
		\centering
		\begin{minipage}[b]{0.31\textwidth}
			\includegraphics[width=1\textwidth, trim={0cm 0cm 0cm 0cm}, clip]{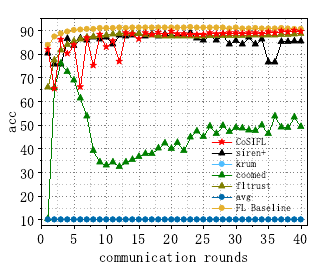}
		\end{minipage}
	}
	\caption{Untargeted attacks, K=10, medium data quality}
	\label{Untargeted attacks, non-iid=0.4}
\end{figure*}

\begin{figure*}[htbp]
	\centering
	\subfigure[sign flipping attack]{
		\centering
		\begin{minipage}[b]{0.31\textwidth}
			\includegraphics[width=1\textwidth, trim={0cm 0cm 0cm 0cm}, clip]{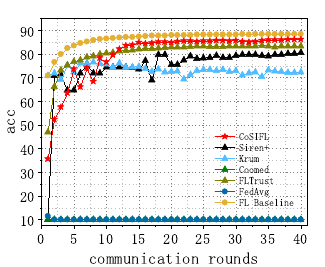}
		\end{minipage}
	}
	\subfigure[label flipping attack]{
		\centering
		\begin{minipage}[b]{0.31\textwidth}
			\includegraphics[width=1\textwidth, trim={0cm 0cm 0cm 0cm}, clip]{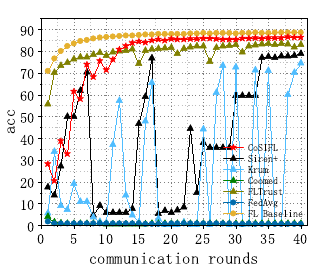}
		\end{minipage}
	}
	\subfigure[adaptive attack]{
		\centering
		\begin{minipage}[b]{0.31\textwidth}
			\includegraphics[width=1\textwidth, trim={0cm 0cm 0cm 0cm}, clip]{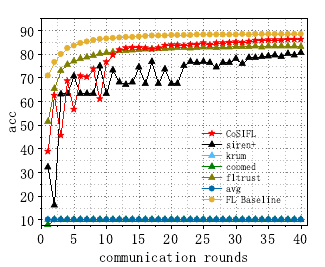}
		\end{minipage}
	}
	\caption{Untargeted attacks, K=50, low data quality}
	\label{Untargeted attacks, non-iid=0.8}
\end{figure*}

\subsection{Effectiveness in defending against targeted attacks}
\begin{figure*}[htbp]
	\centering
	\subfigure[k=10, high data quality]{
		\centering
		\begin{minipage}[b]{0.31\textwidth}
			\includegraphics[width=1\textwidth, trim={0cm 0cm 0cm 0cm}, clip]{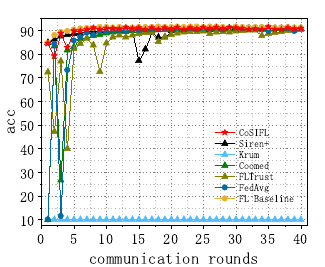}
		\end{minipage}
	}
	\subfigure[k=10, medium data quality]{
		\centering
		\begin{minipage}[b]{0.31\textwidth}
			\includegraphics[width=1\textwidth, trim={0cm 0cm 0cm 0cm}, clip]{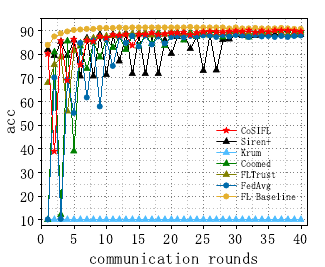}
		\end{minipage}
	}
	\subfigure[k=50, low data quality]{
		\centering
		\begin{minipage}[b]{0.31\textwidth}
			\includegraphics[width=1\textwidth, trim={0cm 0cm 0cm 0cm}, clip]{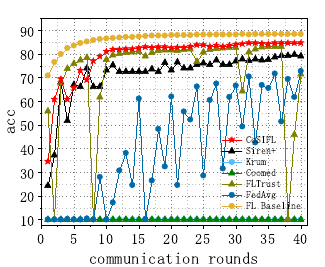}
		\end{minipage}
	}
	\caption{Targeted attack}
	\label{Targeted attacks}
\end{figure*}

Compared to untargeted Byzantine attacks, targeted model poisoning attacks are more challenging because their goal is to only alter specific predictions without affecting the overall performance of the model. In addition to analyzing training curves, we also use "misclassification confidence" to measure the effectiveness of the attack. As shown in Figure \ref{Targeted attacks}, Krum performs the worst in defense. This is because Krum’s defense mechanism assumes that malicious participants work independently and exhibit clear anomalies. However, in targeted attacks, multiple malicious participants may collaborate, bypassing Krum’s defense strategy and leading to suboptimal performance. The remaining methods perform well under both high and medium data quality because targeted attacks only alter specific predictions without impacting the overall model performance. However, their defense effectiveness decreases under low data quality. From Figure~\ref{Targeted attacks,mali}, we can see that only CoSIFL and Siren+ truly defend against targeted model poisoning attacks, as they maintain low misclassification confidence throughout the training process. Under low data quality, CoSIFL exhibits lower misclassification confidence than Siren+, showing better defense performance.

\begin{figure*}[htbp]
	\centering
	\subfigure[k=10, high data quality]{
		\centering
		\begin{minipage}[b]{0.31\textwidth}
			\includegraphics[width=1\textwidth, trim={0cm 0cm 0cm 0cm}, clip]{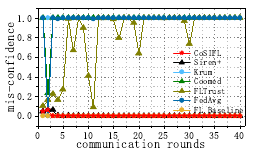}
		\end{minipage}
	}
	\subfigure[k=10, medium data quality]{
		\centering
		\begin{minipage}[b]{0.31\textwidth}
			\includegraphics[width=1\textwidth, trim={0cm 0cm 0cm 0cm}, clip]{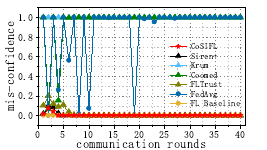}
		\end{minipage}
	}
	\subfigure[k=50, low data quality]{
		\centering
		\begin{minipage}[b]{0.31\textwidth}
			\includegraphics[width=1\textwidth, trim={0cm 0cm 0cm 0cm}, clip]{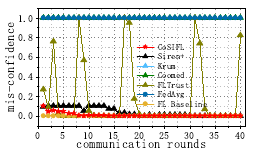}
		\end{minipage}
	}
	\caption{Targeted attack with misclassification confidence}
	\label{Targeted attacks,mali}
\end{figure*}

\subsection{Effectiveness in defending against inference attacks}
\begin{figure*}[htbp]
	\centering
	\subfigure[MLP MIA]{
		\centering
		\begin{minipage}[b]{0.31\textwidth}
			\includegraphics[width=1\textwidth, trim={0cm 0cm 0cm 0cm}, clip]{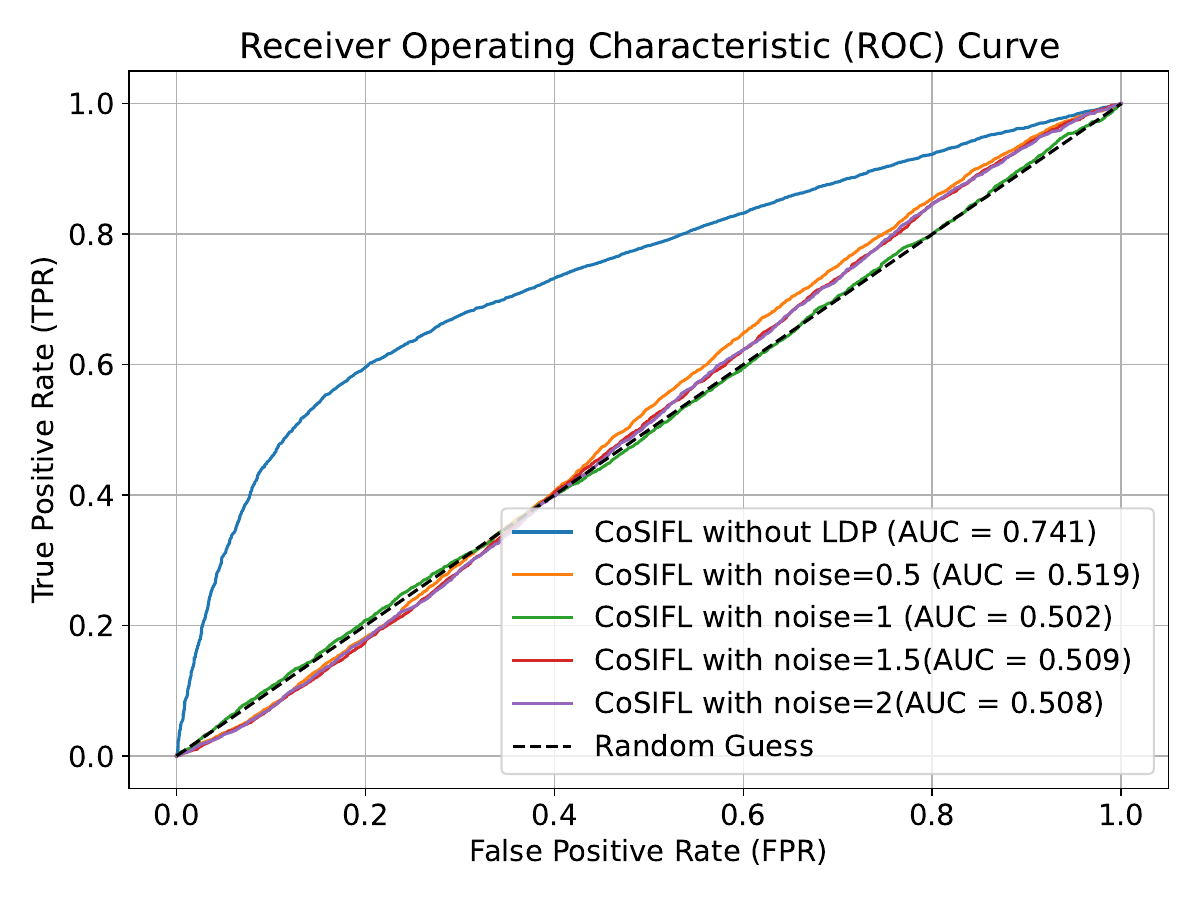}
		\end{minipage}
	}
	\subfigure[Threshold MIA]{
		\centering
		\begin{minipage}[b]{0.31\textwidth}
			\includegraphics[width=1\textwidth, trim={0cm 0cm 0cm 0cm}, clip]{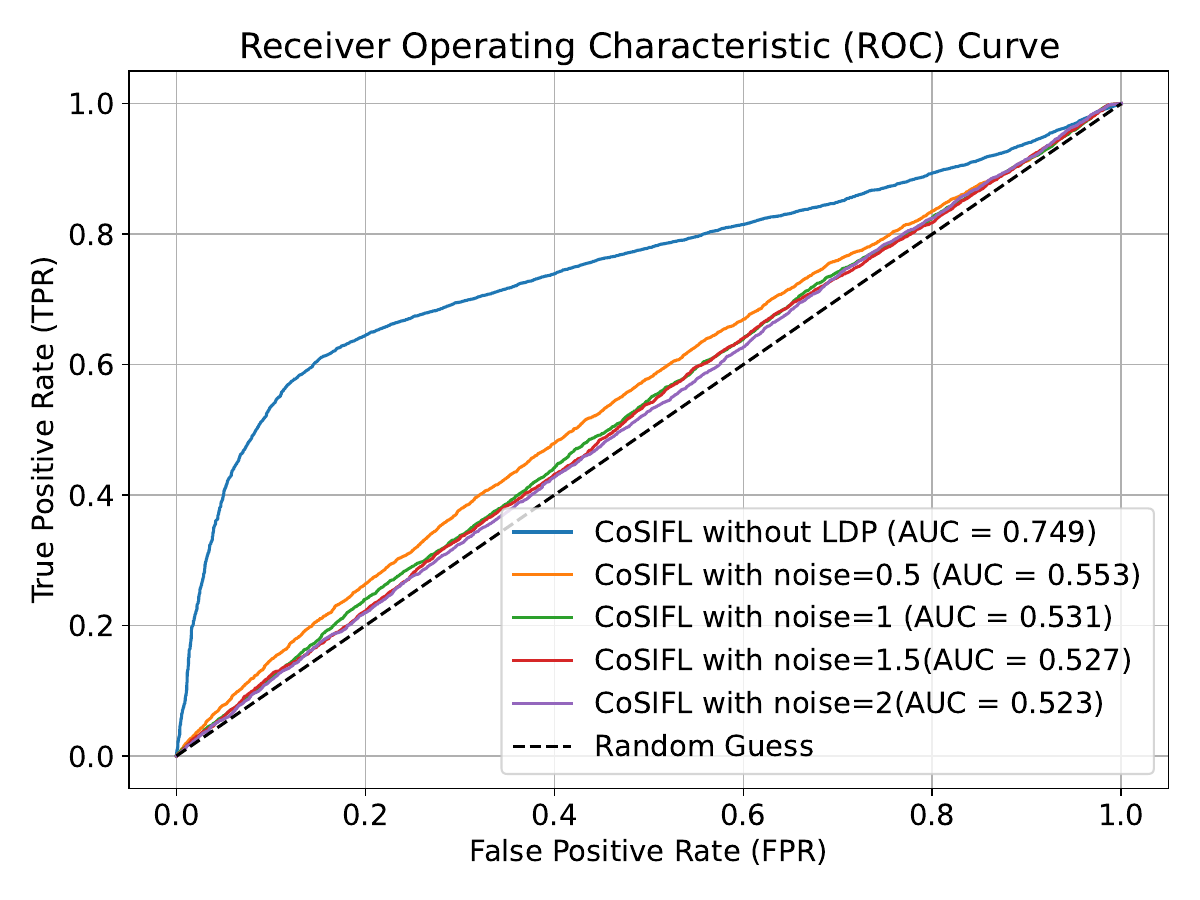}
		\end{minipage}
	}
	\subfigure[LR MIA]{
		\centering
		\begin{minipage}[b]{0.31\textwidth}
			\includegraphics[width=1\textwidth, trim={0cm 0cm 0cm 0cm}, clip]{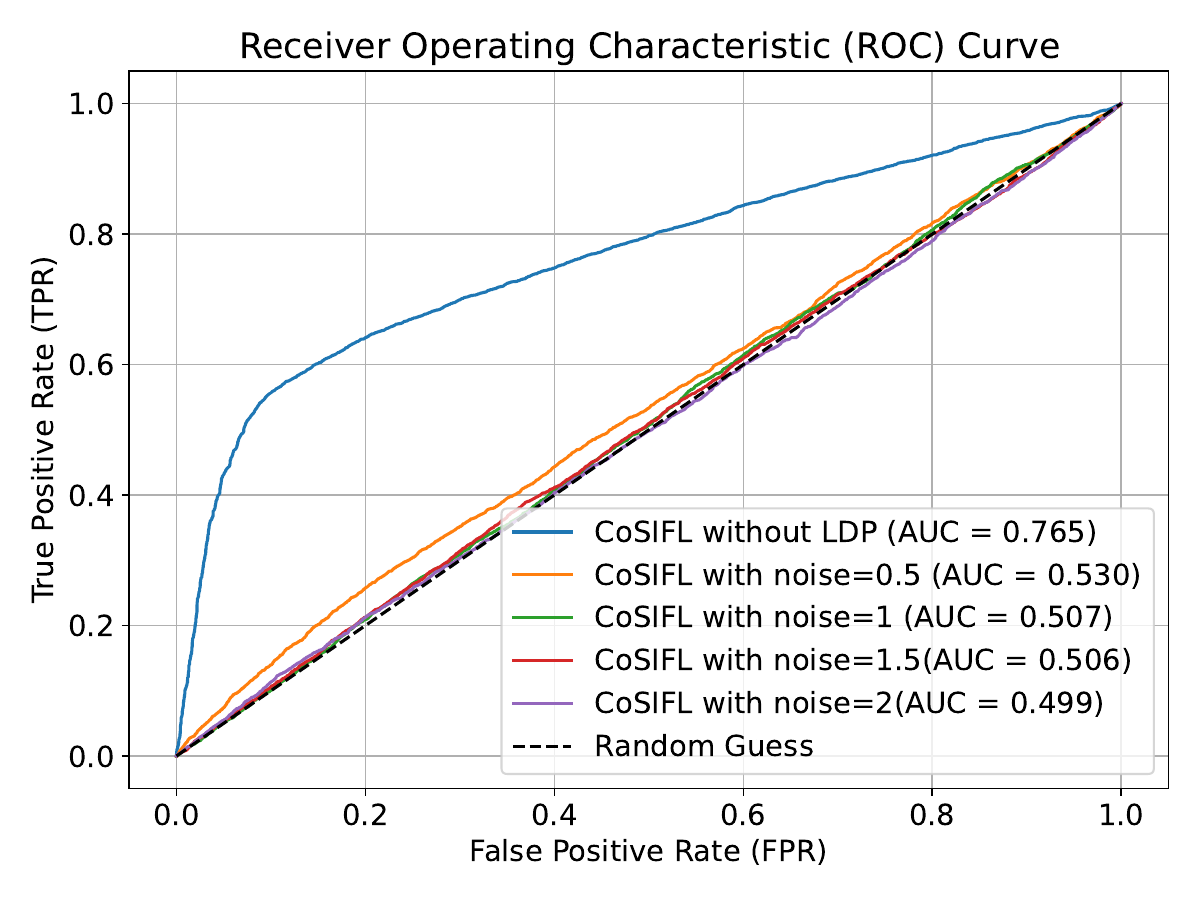}
		\end{minipage}
	}
	\caption{Effectiveness of CoSIFL in defending against MIA.}
	\label{ROC}
\end{figure*}

To evaluate the effectiveness of CoSIFL in defending against inference attacks—particularly Membership Inference Attacks (MIA)—we conduct experiments using three representative attack methods: MLP MIA, Threshold MIA, and LR MIA. As shown in Fig.~\ref{ROC}, the defense performance is measured using ROC curves and their corresponding AUC values. The experiments are conducted on the CIFAR-10 dataset.
Across the three subfigures, we observe the following trends:
Without enabling the Local Differential Privacy (LDP) mechanism, the AUC scores of CoSIFL models are significantly higher than 0.74, indicating a severe privacy leakage risk in the absence of explicit protection mechanisms.
Once LDP is enabled, privacy leakage is substantially mitigated. As the noise intensity increases (from 0.5 to 2.0), the AUC values under all three types of attacks gradually approach the ideal baseline of 0.5, effectively reducing the attacker’s ability to distinguish member from non-member samples. Under LDP noise levels $\geq 1.0$, all three types of attacks are suppressed to near-random guess levels, demonstrating that CoSIFL retains strong defensive capabilities even against advanced inference attacks.

\subsection{Server's cost}

Table~\ref{Server's cost} presents the average server-side cost under different client selection and incentive schemes, including CoSIFL, ND (No Discrimination), NDT (No Discrimination and Training Time), and TPUC (Two-Part Uniform Contract), across various total client counts \( K \). The values in parentheses indicate the percentage increase in cost compared to CoSIFL, i.e., cost reduction (CR).
From the table, it is evident that CoSIFL consistently achieves the lowest server cost, showcasing its efficiency in balancing model performance, security, and incentive overhead. The key observations and analysis are as follows:

CoSIFL achieves the optimal trade-off between client contribution and reward allocation through its Stackelberg game-based design and Pareto-aware client selection. It adaptively selects clients with high data quality and low latency, resulting in minimized training time and reward waste. As \( K \) increases, CoSIFL can better leverage diversity in client attributes, further reducing cost (e.g., from 35,000 at \( K = 10 \) to just 3,509 at \( K = 100 \)).
ND incurs significantly higher cost—up to 82.72\% more than CoSIFL when \( K = 100 \)—because it fails to distinguish between clients with different data quality or privacy settings. Though it considers latency, it uniformly selects clients without regard for their actual utility, leading to suboptimal reward distribution and wasted resources.
NDT performs even worse, as it not only ignores client heterogeneity but also removes time constraints and includes all candidate clients in each round. This results in high latency, more training iterations, and inflated rewards. At \( K = 50 \), its cost is 81.8\% higher than CoSIFL, showing poor scalability under practical constraints.
TPUC attempts to introduce asymmetric reward contracts but lacks fine-grained modeling of client attributes like non-IID degree or alarm accuracy. While more sophisticated than ND/NDT, TPUC still incurs up to 87.61\% higher cost than CoSIFL. This is mainly due to static contract structures and insufficient adaptability to adversarial environments.

In summary, CoSIFL achieves significant cost savings — over 80\% in large-scale settings — by selecting high-quality, trustworthy clients and allocating rewards proportionally based on real contribution and alarm reliability, something none of the other schemes can fully account for. This makes CoSIFL highly suitable for real-world federated learning deployments where cost, security, and participation must be simultaneously optimized.

\begin{table}[h!]
	\centering
	\begin{tabular}{@{}p{0.6cm}p{1cm}p{1.8cm}p{1.8cm}p{1.8cm}@{}}
		\toprule
		K & CoSIFL & ND(CR) & NDT(CR) & TPUC(CR) \\ \midrule
		10 & 35,000 & 50,000(30\%) & 58,350(40.2\%) & 67,668(48.28\%) \\
		30 & 11,676 & 35,003(66.64\%)       & 42,790(72.71\%)       & 50,000(76.65\%)       \\ 
		50 & 7,008       &  30,450(76.99\%)      & 38,511(81.8\%) &  43,333(83.83\%)  \\ 
		80 & 4383       &  22,076(80.14\%) & 28,401(84.57\%)& 31,250(85.97\%)       \\ 
		100 & 3509      &  20,310(82.72\%)      & 26,584(86.8\%)       & 28,333(87.61\%)  \\ 
		\bottomrule
	\end{tabular}
	\caption{Mean server cost across different schemes.}
	\label{Server's cost}
\end{table}

\subsection{Self-Recovery Capability of CoSIFL}

\begin{figure}
	\centering
	\includegraphics[width=0.5\textwidth]{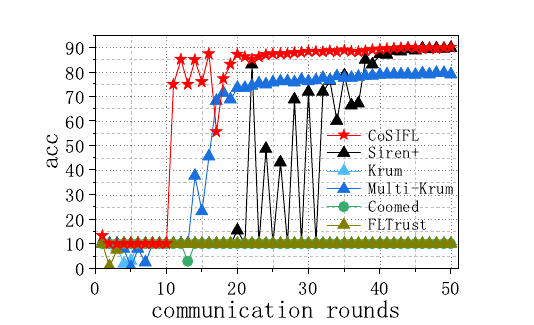}
	\caption{Self-Recovery Capability of CoSIFL, k =10, high data quality, sign flipping attack}
	\label{Recovering Compromised Training}
\end{figure}

To evaluate whether CoSIFL can effectively recover the FL process when the server suffers from early-stage attacks, we design an experiment where no defense mechanisms are enabled during the first 10 communication rounds. From the 11th round onward, the server activates various defense strategies, including CoSIFL and five other baseline methods.
As shown in Fig.~\ref{Recovering Compromised Training}, during the first 10 rounds (with no defense enabled), the global model suffers a significant performance drop due to severe disruption from malicious clients, and all methods struggle to maintain meaningful accuracy. Starting from round 11, CoSIFL exhibits a rapid performance boost, steadily converging from round 22 onward, and reaching around 90\% accuracy by round 50. Siren+ also demonstrates strong recovery ability; however, its convergence is slower than CoSIFL due to the lack of incentive mechanisms for client participation. Multi-Krum shows moderate recovery, benefiting from improved robustness via multiple update averaging, though its recovery pace remains slow.

In contrast, the other methods fail to recover, highlighting CoSIFL’s superior robustness and self-recovery capability. Once the server regains control and reactivates defense mechanisms, CoSIFL effectively filters poisoned updates and restores training reliability and model performance, even after a prolonged attack phase.

\section{My Appendix}

\subsection{Global Model Loss}{\label{global model loss}}

\begin{equation}{\label{model_loss}}
	L = {\gamma _1}{\phi ^T}\theta  + \\
	(1 - {\phi ^T})({\gamma _2}\sum\nolimits_{k \in K} {\frac{{{T^2}}}{{B_k^2\varepsilon _k^2}} + } {\gamma _3}\kappa (\{ {\alpha _k}\} ,\{ {B_k}\} ))
\end{equation}
This expression can be divided into two main parts:
\(\gamma_1 \phi^T \Theta\): \(\Theta = F(w_0) - F(w^*)\) represents the loss gap between the initial model and the optimal model in an ideal (noise-free) condition. \(\phi^T\) characterizes how the global model loss decays, either exponentially or geometrically, as the number of global iterations \(T\) increases (\(\phi = 1 - 2\mu\eta + 2\mu\rho\eta^2\) is a type of decay factor).\(\gamma_1\) signifies the server’s weight on initial convergence speed.\((1-\phi^T)\bigl(\gamma_2 \sum_{k \in N}\frac{T^2}{B_k^2 \,\varepsilon_k^2} + \gamma_3\,\kappa(\{\alpha_k\},\{B_k\})\bigr)\): \((1-\phi^T)\) indicates that when the number of iterations \(T\) becomes large, the initial gradient gap has mostly converged, making other factors (noise and non-IID) the primary concerns. \(\gamma_2 \sum_{k \in N}\tfrac{T^2}{B_k^2 \,\varepsilon_k^2}\) describes the accuracy loss caused by DP noise. When clients have a small \(\varepsilon_k\) (more noise) or the server runs many global iterations \(T\), this term grows larger. \(\gamma_3\,\kappa(\{\alpha_k\},\{B_k\})\) denotes the influence function of non-IID degrees on model accuracy; \(\kappa(\cdot)\) can incorporate factors such as \(\alpha_k\) and batch size \(B_k\) (e.g., a quadratic term reflecting how each client’s share of data deviates from the global distribution). \(\gamma_3\) is its weight.

In summary, formula~\ref{model_loss} aims to integrate: Initial convergence (representing how network or optimization hyperparameters influence convergence rate);Privacy noise(the coupling between \(\varepsilon_k\) and \(T\));Non-IID degree(\(\alpha_k\)).
When the server chooses a client subset \(N\), the number of global iterations \(T\), and an incentive strategy, this loss function \(L(\cdot)\) is incorporated into the overall cost \(C_{\text{server}}\), seeking to maintain the best possible model accuracy while satisfying DP requirements and defending against Byzantine attacks.

\bibliographystyle{unsrtnat}
\bibliography{references}  






\end{document}